\pdfoutput=1

\def\MODE{1} 
\if\MODE1

\else

\fi

\documentclass{article}

\usepackage{graphicx}
\usepackage{soul}
\usepackage{color, xcolor}
\usepackage[thinlines]{easytable}
\usepackage{relsize}
\usepackage{xfrac}
\usepackage{verbatim}
\usepackage{algorithm}
\usepackage[noend]{algorithmic}
\usepackage{amsmath}
\usepackage{amssymb}
\usepackage{amsthm}
\usepackage{epsfig}
\usepackage{wrapfig}
\usepackage{url}
\usepackage[colorlinks=true,citecolor=blue,linkcolor=blue]{hyperref}
\usepackage{multirow}
\usepackage{fullpage}
\usepackage{array}
\usepackage{thmtools} 
\usepackage{thm-restate}

\usepackage[T1]{fontenc}
\usepackage[utf8]{inputenc}
\usepackage{authblk}

\newcommand{\expect}{\mathbb{E}}

\newcommand{\grad}{\nabla}
\newcommand{\real}{\mathbb{R}}
\newcommand{\diag}{\text{diag}}
\newcommand{\eg}{{\it e.g.,} }
\newcommand{\ie}{{\it i.e.,} }
\newcommand\numberthis{\addtocounter{equation}{1}\tag{\theequation}}

\usepackage{times}
\usepackage{nicefrac}
\usepackage{float}
\usepackage{graphicx}
\usepackage{caption}
\usepackage{subcaption}
\usepackage{amsmath}
\usepackage{amsthm}
\usepackage{amsfonts} 
\usepackage{geometry}

\usepackage{hyperref}

\hypersetup{
	colorlinks=true,
	citecolor=blue,
}

\theoremstyle{plain}
\newtheorem{theorem}{Theorem}[section]
\newtheorem{lemma}[theorem]{Lemma}
\newtheorem{cor}[theorem]{Corollary}

\newtheorem{defn}[theorem]{Definition}
\newtheorem{rem}{Remark}[section]
\newtheorem{asm}{Assumption}

\title{Stability and Generalization of Learning Algorithms\\ that Converge to Global Optima}
\author[1]{Zachary Charles}
\author[2]{Dimitris Papailiopoulos}
\affil[1]{Department of Mathematics, University of Wisconsin-Madison}
\affil[2]{Department of Electrical and Computer Engineering, University of Wisconsin-Madison}

\date{\today}

\begin{document}

\maketitle

\begin{abstract}
We establish novel generalization bounds for  learning algorithms that converge to global minima.
We do so by deriving black-box stability results that only depend on the convergence of a learning algorithm and the geometry around the minimizers of the loss function. 
The results are shown for nonconvex loss functions satisfying the Polyak-\L{}ojasiewicz (PL) and the quadratic growth (QG) conditions. We further show that these conditions arise for  some neural networks with linear activations.

We use our black-box results to establish the stability of  optimization algorithms such as  stochastic gradient descent (SGD), gradient descent (GD), randomized coordinate descent (RCD), and the stochastic variance reduced gradient method (SVRG), in both the PL and the strongly convex setting. Our results match or improve state-of-the-art generalization bounds and can easily be extended to similar optimization algorithms. Finally, we show that although our results imply comparable stability for SGD and GD in the PL setting, there exist simple neural networks with multiple local minima where SGD is stable but GD is not.

\end{abstract}

\section{Introduction}

The recent success of training complex models at state-of-the-art accuracy in many common machine learning tasks 
has sparked significant interest and research in algorithmic machine learning.
In practice, not only can these complex deep neural models   yield zero training loss, they can also generalize surprisingly well \cite{zhang2016understanding, lin2016does}.
Although there has been significant recent work in analyzing the training loss performance of several learning algorithms, our theoretical understanding of their generalization properties falls far below what has been observed empirically.

A useful proxy for analyzing the generalization performance of learning algorithms is  that of {\it stability}.
A training algorithm is stable if small changes in the training set result in small differences in the output predictions of the trained model.
In their foundational work, Bousquet and Elisseeff \cite{bousquet1} establish that {\it stability begets generalization}.

While there has been  stability analysis for   empirical risk minimizers \cite{bousquet1, mukherjee2006learning}, there are far fewer results for commonly used iterative learning algorithms. 
In a recent novel work, Hardt et al. \cite{hardt2} establish  stability bounds for SGD, and discuss algorithmic heuristics that provably increase the stability of SGD models.
Unfortunately,
generalizing their techniques to establish stability bounds for other first-order methods
can be a strenuous task. Showing non-trivial stability for more involved algorithms like SVRG \cite{johnson2013accelerating} (even in the convex case), or SGD in more nuanced non-convex setups is far from straightforward. While \cite{hardt2} provides a clean and elegant analysis that shows stability of SGD for non-convex loss  functions, the result requires very small step-sizes. The step-size is small enough that one may require exponentially many steps for provable convergence to an approximate critical point, under standard smoothness assumptions \cite{ghadimi2013stochastic} (see subsection \ref{appendix:sgd_exp_time}). Generally, there seems to be a trade-off between convergence and stability of algorithms. 
In this work we show that under certain geometric assumptions on the loss function around global minima, we can actually leverage the convergence properties of an algorithm to prove that it is stable.

The goal of this work is to provide black-box and easy-to-use stability results for a variety of learning algorithms in non-convex settings.
We show that this is in some cases possible by decoupling the stability of critical points and their proximity to models trained by iterative algorithms.

\paragraph{Our contribution}

We establish that models trained by  algorithms that converge to local minima are stable under the Polyak-\L{}ojasiewicz (PL) and the quadratic growth (QG) conditions \cite{karimi}.
Informally, these conditions assert that the suboptimality of a model is upper bounded by the norm of its gradient and lower bounded by its distance to the closest global minimizer.
As we  see in the following, these conditions are sufficient  for stability and are general enough to yield useful  bounds for a  variety of  settings.

Our results require weaker   conditions compared to the state-of-the art, while recovering several prior stability bounds. For example in \cite{hardt2} the authors require convexity, or strong convexity.
Gonen and Shalev-Shwartz prove the stability of ERMs for nonconvex, but locally strongly convex loss functions obeying strict saddle inequalities \cite{gonen2017fast}. By contrast, we develop comparable stability results for a large class of functions, where no convexity,  local convexity, or saddle point conditions are  imposed. 
We note that although \cite{hardt2} establishes the stability of SGD for smooth non-convex objectives, the stepsize  selection can be prohibitively small for convergence.  In our bounds, we make no assumptions on the hyper-parameters of the algorithms.

We use our black-box results to directly compare the generalization performance of popular first-order methods in general learning setups. 
While direct proofs of stability seem to require a substantial amount of algorithm-specific analysis, our results are derived from known convergence rates of popular algorithms. 
For  strong convexity---a special case of the PL condition---we recover order-wise the stability bounds of Hardt et al. \cite{hardt2}, but for a large family of optimization algorithms  (\eg SGD, GD, SVRG, etc). 
We show that many of these algorithms offer order-wise similar stability as saddle-point avoiding algorithms in nonconvex problems where all local minima are global \cite{gonen2017fast}.
We finally show that while SGD and GD have analogous stability in the convex setting, this breaks down in the non-convex setting. 
We give an explicit example of a simple 1-layer neural network on which SGD is stable but GD is not. 
Such an example was theorized in \cite{hardt2} (\ie  Figure 10 in the aforementioned paper); here we formalize the authors' intuition.
Our results offer yet another indication that SGD trained models can be more generalizable than full-batch GD ones.

Finally, we give examples of some  machine learning scenarios where the PL condition mentioned above holds true.
Adapting techniques from \cite{hardt1}, 
 we show that deep networks with linear activation functions are PL almost everywhere in the parameter space.
Our theory allows us to derive results similar to those in \cite{kawaguchi2016deep} about local/global minimizers in linear neural networks.

\paragraph{Prior work}

	The idea of stability analysis has been around for more than 30 years since the work of Devroye and Wagner \cite{devroye}. Bousquet and Elisseef \cite{bousquet1} defined several notions of algorithmic stability and used them to derive bounds on generalization error. Further work has focused on stability of randomized algorithms \cite{elisseeff} and the interplay between uniform convergence and generalization \cite{shalev}. Mukherjee et al. \cite{mukherjee2006learning}  show that 
	stability implies consistency of empirical risk minimization. Shalev-Shwartz et al. \cite{shalev} show that stability can also imply learnability is some problems.

	Our work is heavily influenced by that of Hardt et al. \cite{hardt2} that establish stability bounds for stochastic gradient descent (SGD) in the convex, strongly convex, and non-convex case.
The	work by Lin et al.~\cite{lin2016generalization} shows that stability of SGD can be controlled by forms of regularization. In \cite{kuzborskij2017data}, the authors give stability bounds for SGD that are data dependent. Since they do not rely on worst-case arguments, they lead to smaller generalization error bounds than that in \cite{hardt2}, but require assumptions on the underlying data.
The	work by Liu et al. \cite{liu2017algorithmic} gives a related notion of {\it uniform hypothesis stability} and show that it implies guarantees on the generalization error. 

	Stability is closely related to the notion of {\it differential privacy} introduced in \cite{dwork2006}. 
	Roughly speaking, differential privacy ensures that the probability of observing any outcome from a statistical query changes  if you modify any single dataset element. 
	Dwork et al. later showed that differentially private algorithms generalize well \cite{dwork2015preserving}. These generalization bounds were later improved by Nissim and Stemmer \cite{nissim2015generalization}. Such generalization bounds are similar to those guaranteed by stability but often require different tools to handle directly.

\section{Preliminaries}\label{section_prelim}

	We first introduce some notation we will use in this paper. For $w \in \real^m$, we will let $||w||$ denote the $2$-norm of $w$. For a matrix $A \in \real^{n\times m}$, we will let $\sigma_{\min}(A)$ denote its minimum singular value, and $||A||_{F}$ denote its Frobenius norm.

	Let $S = \{z_1,\ldots, z_n\}$ be a set of training data, where $z_i \overset{iid}{\sim} \mathcal{D}$. For a model $w$ and a loss function $\ell$, let $\ell(w;z)$ be the error of $w$ on the training example $z$. We define the {\it expected risk} of a model $w$ by
	$R[w] := \expect_{z\sim\mathcal{D}}\ell(w;z)$. Since, we do not have access to the underlying distribution $\mathcal{D}$ optimizing $R[w]$ directly is not possible. 
	Instead, we will measure the {\it empirical risk} of a model $w$ on a set $S$, given by:
	$$R_S[w] := \frac{1}{n}\sum_{i=1}^n \ell(w; z_i).$$ The generalization performance of the model can then be measured by the {\it generalization gap}: 	$$\epsilon_{\text{gen}}(w):=|R_S[w]-R[w]|.$$
	For our purposes, $w$ will be the output of some (potentially randomized) learning algorithm $\mathcal{A}$, trained on some data set $S$. We will denote this output by $\mathcal{A}(S)$. 
	
	Let us now define a related training set 	$S' = \{z_1,\ldots, z_{i-1},z_i',z_{i+1},\ldots, z_n\}$, where $z_i' \sim \mathcal{D}$.  
	We then have the following notion of uniform stability that was first introduced in \cite{bousquet1}.

	\begin{defn}[Uniform Stability] An algorithm $\mathcal{A}$ is uniformly $\epsilon$-stable, if for all data sets $S, S'$ differing in at most one example, we have
	$$\sup_{z} \expect_{\mathcal{A}}\big[\ell(\mathcal{A}(S);z)-\ell(\mathcal{A}(S');z)] \leq \epsilon.$$\end{defn}

	The expectation is taken with respect to the (potential) randomness of the algorithm $\mathcal{A}$. 
	Bousquet and Elisseeff  establish that uniform stability implies small generalization gap \cite{bousquet1}.
	\begin{theorem}\label{thm:gen_error}
	Suppose $\mathcal{A}$ is uniformly $\epsilon$-stable. Then,
	$$|\expect_{S,\mathcal{A}}\big[R_S[\mathcal{A}(S)]-R[\mathcal{A}(S)]\big]| \leq \epsilon.$$
	\end{theorem}

	In practice, uniform stability may be too restrictive, since the bound above must hold for all $z$, irrespective of its marginal distribution. The following notion of stability, while weaker, is still enough to control the generalization gap. Given a data set $S = \{z_1,\ldots, z_n\}$ and $i \in \{1,\ldots,n\}$, we define $S^i$ as $S\backslash z_i$.

	\begin{defn}[Pointwise Hypothesis Stability,  \cite{bousquet1}]$\mathcal{A}$ has pointwise hypothesis stability $\beta$ with respect to a loss function $\ell$ if
	$$\forall i\in\{1,\ldots, n\},\hspace{5pt}\expect_{\mathcal{A},S}\big[| \ell(\mathcal{A}(S);z_i)-\ell(\mathcal{A}(S^i);z_i)| \big] \leq \beta.$$\end{defn}	

	Note that this is a weaker notion than uniform stability, but one can still use it to establish non-trivial generalization bounds:

	\begin{theorem}[\cite{bousquet1}]Suppose we have a learning algorithm $\mathcal{A}$ with pointwise hypothesis stability $\beta$ with respect to a bounded loss function $\ell$ such that $0 \leq \ell(w;z) \leq M$. For any $\delta$, we have with probability at least $1-\delta$,
	$$R[\mathcal{A}(S)] \leq R_{S}[\mathcal{A}(S)] + \sqrt{\dfrac{M^2+12Mn\beta}{2n\delta}}.$$
	\end{theorem}

	In the following, we derive  stability bounds like the above, for models trained on empirical risk functions satisfying the PL and QG conditions. To do so, we will assume that the functions in question are $L$-Lipschitz.

	\begin{defn}[Lipschitz]A function $f: \Omega \to \real$ is $L$-Lipschitz if for all $x_1, x_2 \in \Omega$,
	$$|f(x_1)-f(x_2)| \leq L\|x_1-x_2\|_2.$$\end{defn}

	If $f$ is assumed to be differentiable, this is equivalent to saying that for all $x$, $\|\nabla f(x)\|_2 \leq L$.

	\medskip

	In a recent work, Karimi et al. in \cite{karimi} used the {\it Polyak-\L{}ojasiewicz condition} to prove simplified nearly optimal convergence rates for several first-order methods. Notably, there are some non-convex functions that satisfy the PL condition. The condition is defined below.

	\begin{defn}[Polyak-\L{}ojasiewicz]Fix a set $\mathcal{X}$ and let $f^*$ denote the minimum value of $f$ on $\mathcal{X}$. We will say that a function $f$ satisfies the Polyak-\L{}ojasiewicz (PL) condition on $\mathcal{X}$, if there exists $\mu > 0$ such that for all $x \in \mathcal{X}$ we have

	$$\frac{1}{2}||\nabla f(x)||^2 \geq \mu(f(x)-f^*).$$
	\end{defn}

	Note that for PL functions, every critical point is a global minimizer. 
	While strong convexity implies PL, the reverse is not true. Moreover, PL functions are in general nonconvex (\eg invex functions). 
	We also consider a strictly larger family of functions that satisfy the quadratic growth condition.

	\begin{defn}[Quadratic Growth]We will say that a function $f$ satisfies the quadratic growth (QG) condition on a set $\mathcal{X}$, if there exists $\mu > 0$ such that for all $x \in \mathcal{X}$ we have

	$$f(x)-f^*\ge\frac{\mu}{2}||x_p-x^*||^2,$$
	where $x_p$ denotes the euclidean projection of $x$ onto the set of global minimizers of $f$ in $\mathcal{X}$ (\ie $x_p$ is the closest point to $x$ in $\mathcal{X}$ satisfying $f(x_p) = f^*$).
	\end{defn}

	Both of these conditions have been considered in previous studies. The PL condition was first introduced by Polyak in \cite{lojasiewicz1963topological}, who showed that under this assumption, gradient descent converges linearly. The QG condition has been considered under various guises \cite{bonnans1995second, ioffe1994sensitivity} and can imply important properties about the geometry of critical points. For example, \cite{anitescu2000degenerate} showed that local minima of nonlinear programs satisfying the QG condition are actually isolated stationary points. These kinds of geometric implications will allow us to derive stability results for large classes of algorithms.
\section{Stability of Approximate Global Minima}\label{section_blackbox_results}

	In this section, we establish the stability of large classes of learning algorithms under the PL and QG conditions presented above.
	Our stability results are ``black-box'' in the sense that our bounds are decomposed as a sum of two terms: a term concerning the convergence of the algorithm to a global minimizer, and a term relevant to the geometry of the loss function around the global minima. 
	Both terms are used to establish good generalization and provide some insights into the way that learning algorithms perform.

	For a given data set $S$, suppose we use an algorithm $\mathcal{A}$ to train some model $w$. We let $w_S$ denote the output of our algorithm on $S$. The empirical training error on a data set $S$ is denoted $f_S(w)$ and is given by
	$$f_S(w) = \frac{1}{|S|}\sum_{z \in S} \ell(w;z).$$

	We assume that each of these losses is $L$-Lipschitz with respect to the parameters of the  model. We are interested in conditions on $f_S$ that allow us to make guarantees on the stability of $\mathcal{A}$. As it turns out, the PL and QG condition will allow us to prove such results.
	Although it seems unclear if these conditions are reasonable, in our last section we show that they arise in a large number of machine learning settings, including in certain deep neural networks.

	\subsection{Pointwise Hypothesis Stability for PL/QG Loss Functions}

		To analyze the performance machine learning algorithms, it often suffices to understand the algorithm's behavior with respect to critical points. This requires knowledge of the convergence of the algorithm, and an understanding of the geometric properties of the loss function around critical points. 
		As it turns out, the PL and QG conditions allow us to understand the geometry underlying the minima of our function. Let $\mathcal{X}_{\min}$ denote the set of global minima of $f_S$.

		\begin{theorem}\label{thm:black-box-pl-ptwise}
		Assume that for all $S$ and $w \in \mathcal{X}$, $f_S$ is PL with parameter $\mu$. We assume that applying $\mathcal{A}$ to $f_S$ produces output $w_S$ that is converging to some global minimizer $w_S^*$. Then $\mathcal{A}$ has pointwise hypothesis stability with parameter $\epsilon_{stab}$ satisfying the following conditions.

		{\bf Case 1:} If for all $S$, $\|w_S-w_S^*\| \leq O(\epsilon_{\mathcal{A}})$ then
		$$\epsilon_{stab} \leq O(L\epsilon_{\mathcal{A}}) + \frac{2L^2}{\mu (n-1)}.$$

		{\bf Case 2:} If for all $S$, $|f_S(w_S)-f_S(w_S^*)| \leq O(\epsilon'_{\mathcal{A}})$ then

		$$\epsilon_{stab} \leq O\left(L\sqrt{\frac{\epsilon'_{\mathcal{A}}}{\mu}}\right) + \frac{2L^2}{\mu (n-1)}.$$

		{\bf Case 3:} If for all $S$, $\|\nabla f_S(w_S)\| \leq O(\epsilon''_{\mathcal{A}})$, then

		$$\epsilon_{stab} \leq O\left(\frac{L\epsilon''_{\mathcal{A}}}{\mu}\right) + \frac{2L^2}{\mu (n-1)}.$$

		\end{theorem}

		Suppose our loss functions are PL and our algorithm $\mathcal{A}$ is an oracle that returns a global optimizer $w_S^*$. Then the terms $\epsilon_{\mathcal{A}}, \epsilon'_{\mathcal{A}}, \epsilon''_{\mathcal{A}}$ above are all identical to $0$, leading to the following corollary.

		\begin{cor}\label{cor:pl-bound}
		Let $f_S$ satisfy the PL inequality with parameter $\mu$ and let
		$\mathcal{A}(S) = \text{arg}\min_{w\in\mathcal{X}} f_S(w)$.
		Then, $\mathcal{A}$ has pointwise hypothesis stability with
		$$\epsilon_{stab} = \frac{2L^2}{\mu (n-1)}.$$\end{cor}

		Bousquet and Ellisseef \cite{bousquet1} considered the stability of empirical risk minimizers where the loss function satisfied strong convexity. Their work implies that for $\lambda$-strongly convex functions, the empirical risk minimizer has we stability satisfying
		$\epsilon_{stab} \leq \frac{L^2}{\lambda n}$.
		Since $\lambda$-strongly convex implies $\lambda$-PL, Corollary \ref{cor:pl-bound} generalizes their result, with only a constant factor loss.

		\begin{rem}\label{rem1}
		Theorem~\ref{thm:black-box-pl-ptwise}  holds even if we only have information about $\mathcal{A}$ in expectation. For example, if we only know that $\expect_{\mathcal{A}}\|w_S-w_S^*\| \leq O(\epsilon_{\mathcal{A}})$, we still establish pointwise hypothesis stability  ( in expectation with respect to $\mathcal{A}$),
		with the same constant as above.
		This allows us to apply our result to algorithms such as SGD where we are interested in the convergence in expectation.
		\end{rem}

		A similar result to Theorem~\ref{thm:black-box-pl-ptwise} can be derived for empirical risk functions  satisfy the QG condition and  are {\it realizable}, \eg where zero training loss is achievable.

		\begin{theorem}\label{thm:black-box-qg-ptwise}
		Assume that for all $S$ and $w \in \mathcal{X}$, $f_S$ is QG with parameter $\mu$ and that all its global minima $u$ satisfy $f_S(u) = 0$. We assume that applying $\mathcal{A}$ to $f_S$ produces output $w_S$ that is converging to some global minimizer $w_S^*$. We also assume that for all feasible $w$ and $z$, $|\ell(w;z)| \leq c$. Then $\mathcal{A}$ has pointwise hypothesis stability with parameter $\epsilon_{stab}$ satisfying the following conditions.

		{\bf Case 1:} If for all $S$, $\|w_S-w_S^*\| \leq O(\epsilon_{\mathcal{A}})$ then
		$$\epsilon_{stab} \leq O(L\epsilon_{\mathcal{A}}) + 2L\sqrt{\dfrac{c}{\mu n}}.$$

		{\bf Case 2:} If for all $S$, $|f_S(w_S)-f_S(w_S^*)| \leq O(\epsilon'_{\mathcal{A}})$ then

		$$\epsilon_{stab} \leq O\left(L\sqrt{\frac{\epsilon'_{\mathcal{A}}}{\mu}}\right) + 2L\sqrt{\dfrac{c}{\mu n}}.$$
		\end{theorem}

\begin{rem}
Observe that unlike the case of PL empirical losses, QG empirical losses only allow for a $O(\frac{1}{\sqrt{n}})$ convergence rate of stability.
Moreover, similarly to our result for PL loss functions, the result of Theorem~\ref{thm:black-box-qg-ptwise} holds even if we only have information about the convergence of $\mathcal{A}$ in expectation.
\end{rem}
		
Finally, we can obtain the following Corollary for empirical risk minimizers.
\begin{cor}\label{cor:pl-bound}
		Let $f_S$ satisfy the QG inequality with parameter $\mu$ and let
		$\mathcal{A}(S) = \text{arg}\min_{w\in\mathcal{X}} f_S(w)$.
		Then, $\mathcal{A}$ has pointwise hypothesis stability with
		$$\epsilon_{stab} =2L\sqrt{\dfrac{c}{\mu n}}.$$\end{cor}

	\subsection{Uniform Stability for PL/QG Loss Functions}

Under a more restrictive setup, we can obtain similar bounds for uniform hypothesis stability, which is a stronger stability notion compare to its pointwise hypothesis variant.
The usefulness of uniform stability compared to pointwise stability, is that it can lead to generalization bounds that concentrate exponentially faster \cite{bousquet1} with respect to the sample size $n$.

		As before, given a data set $S$, we let denote $w_S$ be the model that $\mathcal{A}$ outputs. Let $\pi_S(w)$ denote the closest optimal point of $f_S$ to $w$. We will denote $\pi_S(w_S)$ by $w_S^*$. Let $S, S'$ be data sets differing in at most one entry. We will make the following technical assumption: 
\begin{asm}
\label{asm:1}
The empirical risk minimizers for $f_S$ and $f_{S'}$, \ie $w_S^*, w_{S'}^*$ satisfy $\pi_S(w_{S'}^*) = w_S^*$, where $\pi_S(w)$ is the projection of $w$ on the set of empirical risk minimizers of $f_S$. Note that this is satisfied if for every data set $S$, there is a unique minimizer $w^*_S$.
\end{asm}

\begin{rem}
We would like to note that the above assumption is extremely strict, and in general does not apply to empirical losses with infinitely many global minima. To tackle the existence of infinitely many global minima, one could imagine designing $\mathcal{A}(S)$ to output a structured empirical risk minimizer, \eg one such that if $\mathcal{A}$ is applied on $S'$, its projection on the optima of $f_S$ would always yield back $A(S)$. This could be possible, if $A(S)$ corresponded to minimizing instead a regularized, or structure constrained cost function whose set of optimizers only contained a small subset of the global minima of $f_S$.  Unfortunately, coming up with such a structured empirical risk minimizer for general nonconvex losses seems far from straightforward, and serves as an interesting open problem.
\end{rem}	

		\begin{theorem}\label{thm:black-box-pl}
		Assume that for all $S$, $f_S$ satisfies the PL condition with constant $\mu$, and suppose that Assumption~\ref{asm:1} holds. Then $\mathcal{A}$ has uniform stability with parameter $\epsilon_{stab}$ satisfying the following conditions.

		{\bf Case 1:} If for all $S$, $\|w_S-w_S^*\| \leq O(\epsilon_{\mathcal{A}})$ then
		$$\epsilon_{stab} \leq O(L\epsilon_{\mathcal{A}}) + \frac{2L^2}{\mu n}.$$

		{\bf Case 2:} If for all $S$, $|f_S(w_S)-f_S(w_S^*)| \leq O(\epsilon'_{\mathcal{A}})$ then
		$$\epsilon_{stab} \leq O\left(L\sqrt{\frac{\epsilon'_{\mathcal{A}}}{\mu}}\right) + \frac{2L^2}{\mu n}.$$

		{\bf Case 3:} If for all $S$, $\|\nabla f_S(w_S)\| \leq O(\epsilon''_{\mathcal{A}}s)$, then
		$$\epsilon_{stab} \leq O\left(\frac{L\epsilon''_{\mathcal{A}}}{\mu}\right) + \frac{2L^2}{\mu n}.$$
		\end{theorem}

		Since strong convexity is a special case of PL, this theorem implies that if we run enough iterations of a convergent algorithm $\mathcal{A}$ on a $\lambda$-strongly convex loss function, then we would expect uniform stability on the order of
		$$\epsilon_{stab} = O\bigg(\frac{L^2}{\lambda n}\bigg).$$ In particular, this theorem recovers the stability estimates for ERMs and SGD applied to strongly convex functions proved in \cite{bousquet1} and \cite{hardt2}, respectively.

		In order to make this result more generally applicable, we would like to extend the theorem to a larger class of functions than just globally PL functions. If we assume boundedness of the loss function, then we can derive a similar result for globally QG functions. This leads us to the following theorem:
		\begin{theorem}\label{thm:black-box-qg}
		Assume that for all $S$, $f_S$ satisfies the QG condition with parameter $\mu$, moreover let Assumption~\ref{asm:1} hold. Suppose that for all $z$ and $w \in \mathcal{X}$, $\ell(w;z) \leq c$. Then $\mathcal{A}$ is uniformly stable with parameter $\epsilon_{stab}$ satisfying:

		{\bf Case 1:} If for all $S$, $\|w_S-w_S^*\| \leq O(\epsilon_{\mathcal{A}})$ then for all $z$ we have:
		$$\epsilon_{stab} \leq O(L\epsilon_{\mathcal{A}}) + 2L\sqrt{\frac{c}{\mu n}}.$$

		{\bf Case 2:} If for all $S$, $|f_S(w_S)-f_S^*| \leq O(\epsilon'_{\mathcal{A}})$ then for all $z$ we have:
		$$\epsilon_{stab}\leq O\left(L\sqrt{\frac{\epsilon'_{\mathcal{A}}}{\mu}}\right) + 2L\sqrt{\frac{c}{\mu n}}.$$

		\end{theorem}

		\begin{rem}By analogous reasoning to that in Remark \ref{rem1}, both Theorem \ref{thm:black-box-pl} and Theorem \ref{thm:black-box-qg} hold if you only have information about the output of $\mathcal{A}$ in expectation.\end{rem}
\section{PL loss functions in practice}\label{section_pl_functions}

	\subsection{Strongly Convex Composed with Piecewise-Linear Functions}

		As the bounds above show, the PL and QG conditions are sufficient for algorithmic stability and therefore imply good generalization.
		 In this section, we show that the PL condition actually arises in some interesting machine learning setups, including least squares minimization, strongly convex functions composed with piecewise linear functions, and  neural networks with linear activation functions. A first step towards a characterization of PL loss functions was proved by Karimi et al. \cite{karimi}, which established that the composition of a strongly-convex function and a linear function results in a loss that satisfies the PL condition.

		We wish to generalize this result to piecewise linear activation functions. Suppose  that $\sigma: \real \to \real$ is defined by $\sigma(z) = c_1z$, for $z>0$ and $\sigma(z) = c_2z, for z\le0$. Here $c_i > 0$. For a vector $z \in \real^n$, we denote by $\sigma(z) \in \real^n$ the vector whose $i$th component is $\sigma(z_i)$. Note that this encompasses leaky-ReLU functions. Following similar techniques to those in \cite{karimi}, we get the following result showing that the composition of strongly convex functions with piecewise-linear functions are PL. The proof can be found in Appendix \ref{proof:str-cvx-relu}.

		\begin{theorem}\label{thm:str-cvx-ReLU}
		Let $g$ be strongly-convex with parameter $\lambda$, $\sigma$ a leaky ReLU activation function with slopes $c1$ and $c_2$, and $X$ a matrix with minimum singular value $\sigma_{\min}(X)$. Let $c = \min\{|c_1|,|c_2|\}$. Then $f(w) = g(\sigma(Xw))$ is PL almost everywhere with parameter $\mu = \lambda\sigma_{\min}(X)^2c^2$.\end{theorem}

		In particular, 1-layer neural networks with a squared error loss and leaky ReLU activations satisfy the PL condition. More generally, this holds for any piecewise-linear activation function with slopes $\{c_i\}_{i=1}^k$. As long as each slope is non-zero and $X$ is full rank, the result above shows that the PL condition is satisfied.

	\subsection{Linear Neural Networks}

		The results above only concern one layer neural networks. Given the prevalence of deep networks, we would like to say something about the associated loss function. As it turns out, we can prove that a PL inequality holds in large regions of the parameter space for deep linear networks.

		Say we are given a training set $S = \{z_1,\ldots, z_n\}$ where $z_i = (x_i,y_i)$ for $x_i, y_i \in \real^d$. Our neural network will have $\ell$ fully-connected non-input layers, each with $d$ neurons and linear activation functions. We will parametrize the neural network model via $W_1,\ldots, W_\ell$, where each $W_i \in \real^{d\times d}$. That is, the output at the first non-input layer is $u_1 = W_1x$ and the output at layer $k \geq 2$ is $A_k u_{k-1}$. Letting $X, Y \in \real^{d\times N}$ be the matrices with $x_i, y_i$ as their columns (respectively), we can then write our loss function as
		$$f(W) = \frac{1}{2}\|W_\ell W_{\ell-1}\ldots W_1X -Y\|_{F}^2.$$

		Let $W = W_\ell W_{\ell-1}\ldots W_1$. The optimal value of $W$ is $W^* = YX^+$. Here, $X^+ = X^T(XX^T)^{-1}$ is the pseudoinverse of $X$. We assume that $X \in \real^{d\times N}$ has rank $d$ so that $XX^T$ is invertible. We will also make use of the following lemma which we prove in Appendix \ref{proof:proj-bound}.

		\begin{lemma}\label{lem:proj-bound}
		Let $W \in \real^{d\times d}$ be some weight matrix. Then for $C = \|(XX^T)^{-1}X\|_{F}^2$, we have
		$$C\|(WX-Y)X^T\|_{F}^2 \geq \|WX-Y\|_{F}^2-\|YX^+X-Y\|_{F}^2.$$
		\end{lemma}

		For a matrix $A$, let $\sigma_{\min}(A)$ denote the smallest singular value of $A$. For a given $W_1,\ldots, W_\ell$, let $W = W_\ell W_{\ell-1}\ldots W_1$. In Appendix \ref{proof:linear-grad-bound}, we prove the following lemma.

		\begin{lemma}\label{lem:linear-grad-bound}
		Suppose that the $W_i$ satisfy $\sigma_{\min}(W_i) \geq \tau > 0$ for all $i$. Then,
		$$\|\nabla f(W_1,\ldots, W_\ell)\|_{F}^2 \geq \ell \tau^{2\ell-2}\|(WX-Y)X^T\|_{F}^2.$$
		\end{lemma}		

		Combining Lemmas \ref{lem:proj-bound} and \ref{lem:linear-grad-bound}, we derive the following interesting corollary about when critical points are global minimizers. This result is not directly related to the work above, but gives an easy way to understand the landscape of critical points of deep linear networks.

		\begin{theorem}\label{thm:local-global-lnn}
			Let $(W_1,\ldots, W_\ell)$ be a critical point such that each $W_i$ has full rank. Then $(W_1,\ldots, W_\ell)$ is a global minimizer of $f$.\end{theorem}
			\begin{proof}
			Since each $W_i$ has full rank, we know that $\tau = \min_{i}\sigma_{\min}(W_i) > 0$. Using Lemma \ref{lem:linear-grad-bound} and the fact that $\nabla f(W_1,\ldots, W_\ell) = 0$, we get $\|(WX-Y)X^T\|_{F}^2 = 0$. Therefore, $WXX^T = YX^T$. Assuming that $(XX^T)$ is invertible, we find that $W = YX^+$, which equals $W^*$.
		\end{proof}

		Thematically similar results have been derived previously for 1 layer networks in \cite{xie2017diverse} and for deep neural networks in \cite{zhou2017landscape}. In \cite{kawaguchi2016deep}, Kawaguchi derives a similar result to ours for deep linear neural networks. Kawaguchi shows that every critical point is either a global minima or a saddle point. Our result, by contrast, implies that all full-rank critical points are global minima.

		Lemmas \ref{lem:proj-bound} and \ref{lem:linear-grad-bound} can also be combined to show that linear networks satisfy the PL condition in large regions of parameter space, as the follwing theorem says.

		\begin{theorem}\label{thm:pl-lnn}
		Suppose our weight matrices $(W_1,\ldots, W_\ell)$ satisfy $\sigma_{\min}(W_i) \geq \tau$ for $\tau > 0$. Then $f(W_1,\ldots, W_\ell)$ satisfies the following PL inequality:

		$$\frac{1}{2}\bigg|\bigg|\dfrac{\partial f}{\partial (W_1,\ldots, W_\ell)}\bigg|\bigg|_{F}^2 \geq  \dfrac{\ell \tau^{2\ell-2}}{\|(XX^T)^{-1}X\|_{F}^2}(f(W_1,\ldots, W_\ell)-f^*).$$
		\end{theorem}

		\begin{proof}
			By Lemma \ref{lem:linear-grad-bound} and Lemma \ref{lem:proj-bound} we find:
			\begin{align*}
			\frac{1}{2}\bigg|\bigg|\dfrac{\partial f}{\partial (W_1,\ldots, W_\ell)}\bigg|\bigg|_{F}^2 &\geq \ell \tau^{2\ell-2}\frac{1}{2}\|(WX-Y)X^T\|_{F}^2\\
			&\geq \ell \tau^{2\ell-2}\dfrac{1}{\|(XX^T)^{-1}X\|_{F}^2}\frac{1}{2}(\|WX-Y\|_{F}^2-\|YX^+X-Y\|_{F}^2)\\
			&= \dfrac{\ell \tau^{2\ell-2}}{\|(XX^T)^{-1}X\|_{F}^2}(f(W_1,\ldots, W_\ell)-f^*).\end{align*}
		\end{proof}		

		In other words, taking
		$$\mu =  \dfrac{\ell \tau^{2\ell-2}}{\|(XX^T)^{-1}X\|_{F}^2},$$
		then at every point $(W_1,\ldots, W_\ell)$ satisfying $\sigma_{\min}(W_i) \geq \tau$, the loss function of our linear network satisfies the PL condition with parameter $\mu$.

\section{Stability of Some First-order Methods}\label{section_algorithms}

		We wish to apply our bounds from the previous section to popular convergent gradient-based methods.
		 We consider SGD, GD, RCD, and SVRG. When we have $L$-Lipschitz, $\mu$-PL loss functions $f_S$ and $n$ training examples, Theorem \ref{thm:black-box-pl} states that any learning algorithm $\mathcal{A}$ has uniform stability $\epsilon_{stab}$ satisfying

		$$\epsilon_{stab} \leq O\bigg(L\sqrt{\dfrac{\epsilon_{\mathcal{A}}}{\mu}}\bigg) + O\bigg(\dfrac{L^2}{\mu n}\bigg).$$

		Here, $\epsilon_{\mathcal{A}}$ refers to how quickly $\mathcal{A}$ converges to the optimal value of the loss function. Specifically, this holds if the algorithm produces a model $w_S$ satisfying $|f_S(w_s)-f_S^*| \leq O(\epsilon_{\mathcal{A}})$. 
		For example, if we want to guarantee that our algorithm has the same stability as SGD in the strongly convex case, then we need to determine how many iterations $T$ we need to perform such that
		\begin{equation}\label{conv_rate}
		\epsilon_{\mathcal{A}} = O\bigg(\dfrac{L^2}{\mu n}\bigg).\end{equation}

 		The convergence rates of SGD, GD, RCD, and SVRG have been studied extensively in the literature \cite{bubeck2015convex, johnson2013accelerating, karimi, nesterov2012efficiency, nesterov2013introductory}. The results are given below. When necessary to state the result, we assume a constant step-size of $\gamma$. Figure \ref{fig1} below summarizes the values of $\epsilon_\mathcal{A}$ for $T$ iterations of SGD, GD, RCD, and SVRG applied to $\lambda$-strongly convex loss functions and $\mu$-PL loss functions. Note that if Eq.~\eqref{conv_rate} holds, then Corollary \ref{cor:pl-bound} implies that our algorithm is uniformly stable with parameter $\epsilon_{stab} = O(L^2/\mu n)$. Moreover, this is the same stability as that of SGD for strongly-convex functions \cite{hardt2}, and saddle point avoiding algorithms on strict-saddle loss functions \cite{gonen2017fast}. 	

		\begin{figure}[H]
			\centering
			\begin{tabular}{ |c|c|c| } 
			\hline
			 & $\lambda$-SC & $\mu$-PL \\ 
			\hline
			& & \\
			SGD & {\Large $O$}$\left((1-2\gamma\lambda)^T + \dfrac{\gamma L^2}{2\lambda}\right)$~\cite{bubeck2015convex} & {\Large $O$}$\left((1-2\gamma\mu)^T + \dfrac{\gamma L^2}{2\mu}\right)$~\cite{karimi} \\ 
			& & \\
			\hline
			& & \\
			GD & {\Large $O$}$\left( \left(1-\dfrac{\lambda}{L}\right)^T\right)$~\cite{nesterov2013introductory} & {\Large $O$}$\left( \left(1-\dfrac{\mu}{L}\right)^T\right)$~\cite{karimi} \\
			& & \\
			\hline
			& & \\
			RCD & {\Large $O$}$\left( \left(1-\dfrac{\lambda}{dL}\right)^T\right)$~\cite{nesterov2012efficiency} & {\Large $O$}$\left( \left(1-\dfrac{\mu}{dL}\right)^T\right)$~\cite{karimi} \\ 
			& & \\
			\hline
			& & \\
			SVRG & {\Large $O$}$\left(\left(\dfrac{1}{\lambda \gamma (1-2L\gamma)m} + \dfrac{2L\gamma}{1-2L\gamma}\right)^T\right)$~\cite{johnson2013accelerating} & {\Large $O$}$\left(\left(\dfrac{1}{\mu \gamma (1-2L\gamma)m} + \dfrac{2L\gamma}{1-2L\gamma}\right)^T\right)$~\cite{karimi} \\
			& & \\
			\hline
			\end{tabular}
			\caption{Convergence rates for $T$ iterations of various gradient-based algorithms in the $\lambda$-SC and $\mu$-PL settings.}
			\label{fig1}
		\end{figure}

		We use the above convergence rates of these algorithms in the $\lambda$-strongly convex and $\mu$-PL settings to determine how many iterations are required such that we get stability that is $O(L^2/\mu n)$. The results are summarized in Figure \ref{fig2} below.

		Note that in the $\mu$-PL case, although it is a nonconvex setup the above algorithms all exhibit the same stability for these values of $T$. 
		This  is not  the case in general: several studies have observed that small-batch SGD offers superior generalization performance compared to large-batch SGD, or full-batch GD, when training  deep neural networks \cite{keskar2016large}.
		
		Unfortunately our bounds above, are not nuanced enough to capture the difference in generalization performance between mini-batch and large-batch SGD.
		Below, we will make this observation formal. Although SGD and GD can be equally stable for nonconvex problems satisfying the PL condition, there exist nonconvex problems where full-batch GD is not stable and SGD is stable.
		\begin{figure}[H]
			\centering
			{\small
			\begin{tabular}{ |c|c|c| } 
			\hline
			 & $\lambda$-SC & $\mu$-PL \\ 
			\hline
			& & \\
			SGD & {\Large $O$}$\left(\dfrac{Ln}{\lambda}\right)$ & {\Large $O$}$\left(\dfrac{Ln}{\mu}\right)$ \\ 
			& & \\
			\hline
			& & \\
			GD & {\Large $O$}$\left( \dfrac{ \log\left(\frac{L}{\lambda n}\right) }{ \log\left(1-\frac{\lambda}{L}\right) } \right)$ & {\Large $O$}$\left( \dfrac{ \log\left(\frac{L}{\mu n}\right) }{ \log\left(1-\frac{\mu}{L}\right) } \right)$ \\ 
			& & \\
			\hline
			& & \\
			RCD & {\Large $O$}$\left( \dfrac{ \log\left(\frac{L}{\lambda n}\right) }{ \log\left(1-\frac{\lambda}{dL}\right) } \right)$ & {\Large $O$}$\left( \dfrac{ \log\left(\frac{L}{\mu n}\right) }{ \log\left(1-\frac{\mu}{dL}\right) } \right)$ \\ 
			& & \\
			\hline
			& & \\
			SVRG & {\Large $O$}$\left( \dfrac{ \log\left(\frac{L}{\lambda n}\right) }{ \log\left(\frac{1}{\lambda \gamma (1-2L\gamma)m} + \frac{2L\gamma}{1-2L\gamma}\right) } \right)$ & {\Large $O$}$\left( \dfrac{ \log\left(\frac{L}{\mu n}\right) }{ \log\left(\frac{1}{\mu \gamma (1-2L\gamma)m} + \frac{2L\gamma}{1-2L\gamma}\right) } \right)$ \\ 
			& & \\
			\hline
			\end{tabular}
			}
			\caption{The number of iterations $T$ that achieves stability $\epsilon_{stab} = O(L^2/\mu n)$ for various gradient-based algorithms with step size $\gamma$ in the $\lambda$-SC and $\mu$-PL settings.}
			\label{fig2}
		\end{figure}	
	\section{The Instability of Gradient Descent}

		In \cite{hardt2}, Hardt et al. proved bounds on the uniform stability of SGD. They also noted that GD does not appear to be provably as stable for the nonconvex case and sketched a situation in which this difference would appear. Due to the similarity of SGD and GD, one may expect similar uniform stability. 
		While this is true in every convex setting  as we show in the Appendix, subsection \ref{sec:gd_stab} (without requiring strong convexity), this breaks down in the non-convex setting. Below we construct an explicit example where GD is not uniformly stable, but SGD is. This example formalizes the intuition given in \cite{hardt2}.

		For $x, w \in \real^m$ and $y \in \real$, we let 
		$$\ell(w; (x,y)) = (\langle w,x\rangle^2 + \langle w,x\rangle - y)^2.$$
		Intuitively, this is a generalized quadratic model where the predicted label $\hat{y}$ for a given $x$ is given by
		$$\hat{y} = \langle w,x\rangle^2 + \langle w,x\rangle.$$
		The above predictive model can be described by the following 1-layer neural network using a quadratic and a linear activation function, denoted $z^2$ and $z$.

		\begin{figure}[H]
			\centering
			\includegraphics[width=0.4\linewidth]{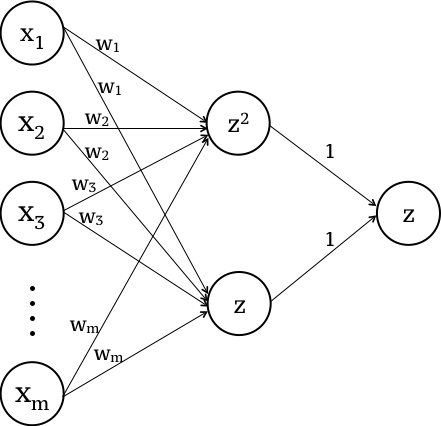}
			\caption{{\small A neural network representing a generalized quadratic model.}}
		\end{figure}

		Note that for a fixed $x, y$, $\ell(w; (x,y))$ is non-convex, \ie it is a quartic polynomial in the weight vector $w$. 
		We will use this function and construct data sets $S, S'$ that differ in only one entry, for which GD produces significantly different models. 
		We consider this loss function for all $z = (x,y)$ with $\|z\| \leq C$ for $C$ sufficiently large. When $m = 1$, the loss function simplifies to
		$$\ell(w; (x,y)) = (w^2x^2+wx-y)^2.$$
		Consider $\ell(w; (x,y))$ at $(-1,1)$ and $(\frac{-1}{2},1)$. Their graphs are as follows. We also graph $g(w)$ which we define as
		$$g(w) = \frac{1}{2}\bigg(\ell\big(w; (-1,1)\big) + \ell\big(w; (\frac{-1}{2},1)\big)\bigg).$$
		\begin{figure}[H]
				\vspace{-0.3cm}
		\begin{center}
		\begin{minipage}{.33\textwidth}
		  \centering
		  \includegraphics[width=1.05\linewidth]{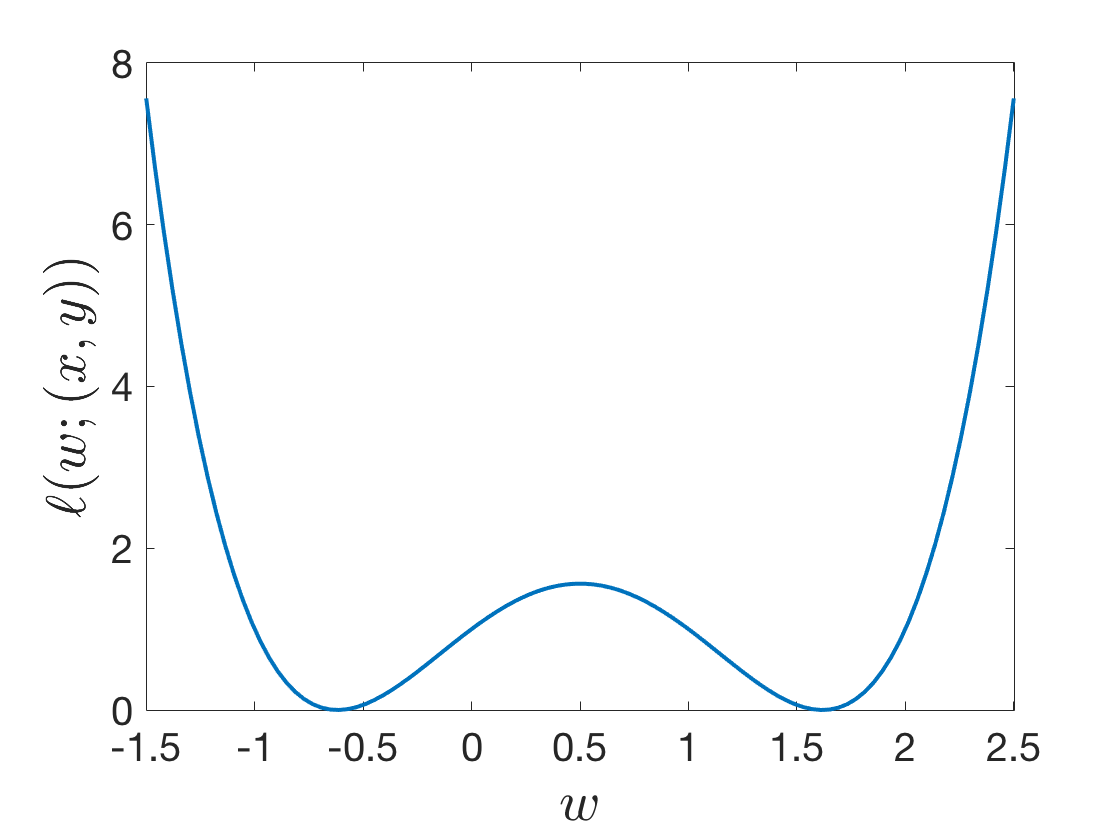}
		  \label{fig:loss1}
		\end{minipage}%
		\begin{minipage}{.33\textwidth}
		  \centering
		  \includegraphics[width=1.05\linewidth]{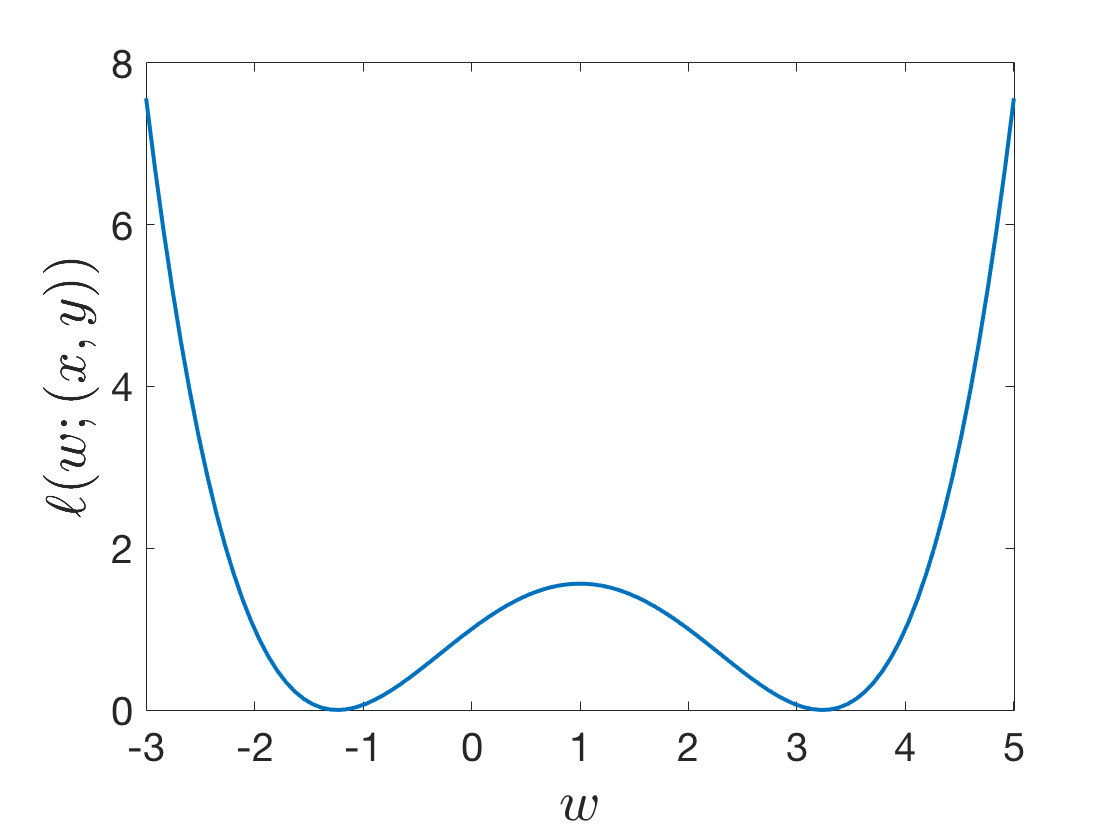}
		  \label{fig:loss2}
		\end{minipage}
		\begin{minipage}{.33\textwidth}
			\centering
			\includegraphics[width=1.05\linewidth]{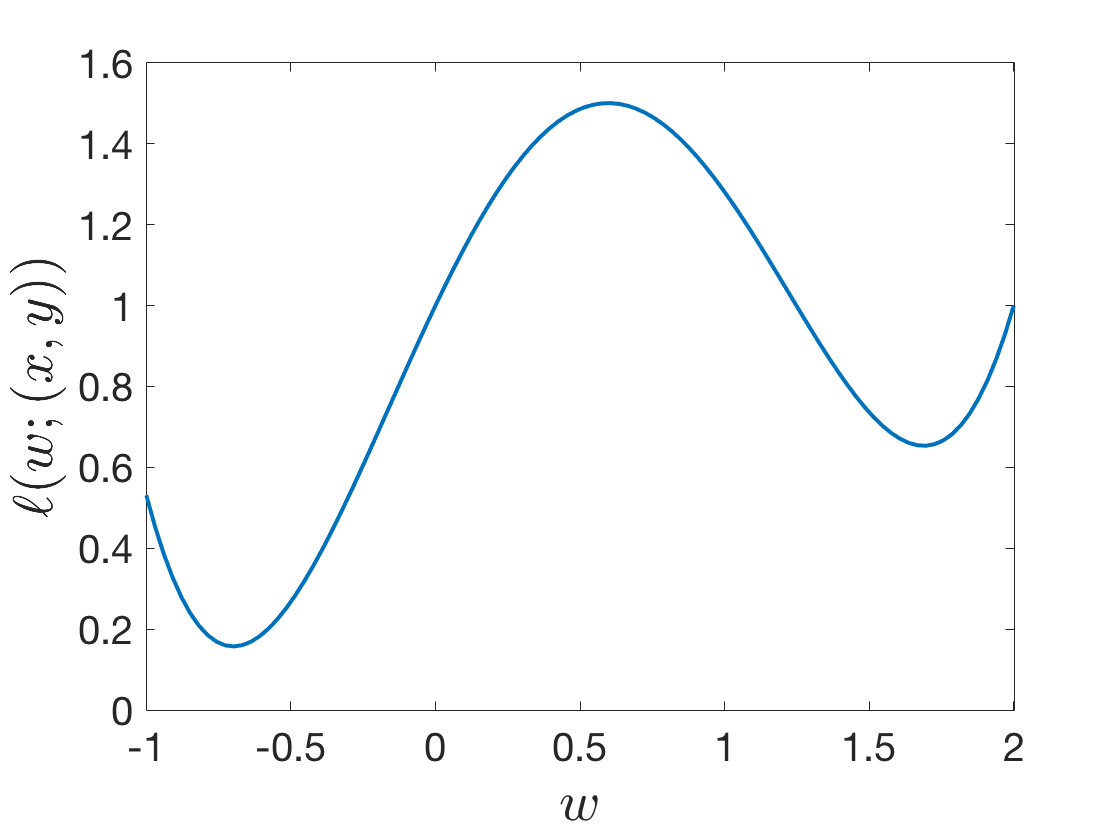}
			\label{fig:loss3}		
		\end{minipage}%
				\end{center}		\vspace{-0.7cm}
		\caption{{\small Graphs of the functions  $\ell(w; (-1,1))$ (left), $\ell(w; (\frac{-1}{2},1))$ (middle), and $g(w) = \frac{1}{2}[\ell(w; (-1,1)) + \ell(w; (\frac{-1}{2},1))]$ (right).}}
		\label{quad_loss}
		\vspace{-0.3cm}
		\end{figure}
		Note that the last function has two distinct basins of different heights. Taking the gradient, one can show that the right-most function in Figure \ref{quad_loss} has zero slope at $\hat{w} \approx 0.598004$. Comparing $\ell(w; (-1,1))$ and $\ell(w; (\frac{-1}{2},1))$, we see that the sign of their slopes agrees on $(-\frac{1}{2}, \frac{1}{2})$ and on $(1,\frac{3}{2})$. The slopes are of different sign in the interval $[\frac{1}{2},1]$. We will use this to our advantage in showing that gradient descent is not stable, while that SGD is.

		We will construct points $(x_1,y_1)$ and $(x_2, y_2)$ such that $\ell(w;(x_1,y_1))$ and $\ell(w;(x_2,y_2))$ have positive and negative slope at $\hat{w}$. To do so, we will first construct an example with a slope of zero at $\hat{w}$. A straightforward computation shows that $\ell(w;(\frac{-1}{2\hat{w}},0))$ has slope zero at $\hat{w}$. A graph of this loss function is given below. Note that $\hat{w}$ corresponds to the concave-down critical point in between the two global minima.

		\begin{figure}[H]
		\centering
		\begin{minipage}{.33\textwidth}
		  \centering
		  \includegraphics[width=1.05\linewidth]{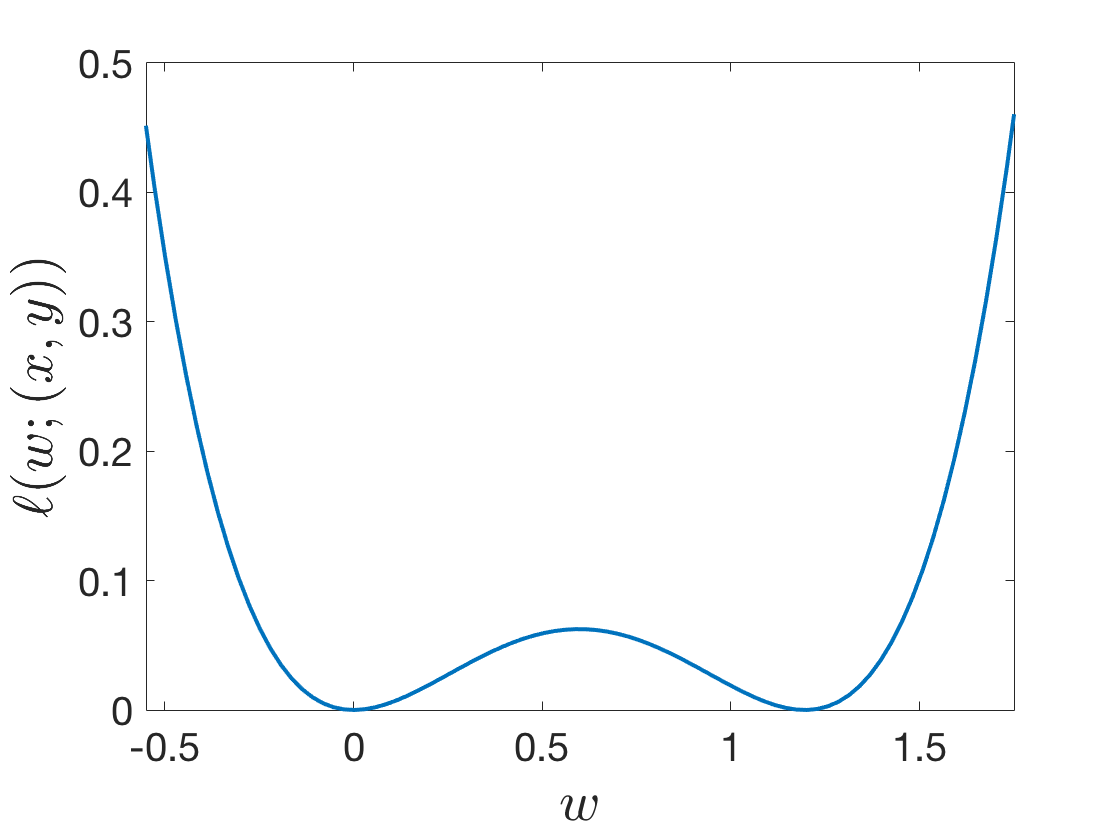}
		  \label{fig:loss4}
		\end{minipage}%
		\caption{{\small Graph of the function $\ell(w;(-\frac{1}{2\hat{w}}),0)$. By construction, this function has critical points at $w = 0, \hat{w}, 2\hat{w}$.}}
		\end{figure}
		Define
		$$z_{\pm} = \left(\frac{-1}{2(\hat{w}\pm\epsilon)},0\right).$$
		Straightforward calculations show that $\ell(w; (z_+,0))$ will have positive slope for $w \in (0,\hat{w}+\epsilon)$, while $\ell(w;(z_-,0))$ will have negative slope for $w \in (\hat{w}-\epsilon,2(\hat{w}-\epsilon))$. In particular, their slopes have opposite signs in the interval $(\hat{w}-\epsilon, \hat{w}+\epsilon)$. Define 
		\begin{gather*}
		S = \{z_1,\ldots, z_{n-1}, z_-\}\\
		S' = \{z_1,\ldots, z_{n-1},z_+\},\end{gather*}
		where $z_i = (-1,1)$ for $1 \leq i \leq \frac{n-1}{2}$, $z_i = (-1/2,1)$ for $\frac{n-1}{2} < i \leq n-1$. By construction, we have
		$$f_S(w) = \frac{n-1}{n}g(w) + \ell\left(w;\left(\frac{-1}{2(\hat{w}+\epsilon)},0\right)\right).$$
		$$f_{S'}(w) = \frac{n-1}{n}g(w) + \ell\left(w;\left(\frac{-1}{2(\hat{w}-\epsilon)},0\right)\right).$$
		Then $f_S(w)$, $f_{S'}(w)$ will approximately have the shape of the right-most function in Figure \ref{quad_loss} above. However, recall that $\frac{d }{d w}g(w) = 0$ at $w = \hat{w}$. Therefore, there is some $\delta$, with $0 < \delta < \epsilon$, such that for all $w \in (\hat{w}-\delta,\hat{w}+\delta)$,
		\begin{equation}\label{sign_eq}
		\dfrac{d}{dw}f_{S}(w) < 0 < \dfrac{d}{dw} f_{S'}(w).\end{equation}
		Now say that we initialize gradient descent with some step-size $\gamma > 0$ in the interval $(w-\delta, w+\delta)$. By \eqref{sign_eq}, the first step of gradient descent on $f_S$ will produce a step moving to the left, but a step moving to the right for $f_{S'}$. This will hold for all $\gamma > 0$. Moreover, the iterations for $f_S$ will continue to move to towards the left basin, while the iterations for $f_{S'}$ will continue to move to the right basin since $\ell(w;(z_+,0))$ has positive slope for $w \in (0,\hat{w}+\epsilon)$ while $\ell(w;(z_-,0))$ has negative slope for $w \in (\hat{w}-\epsilon,2(\hat{w}-\epsilon)$, and that $g(w)$ has positive slope for $w \in (-\frac{1}{2},\hat{w})$ and negative slope for $w \in (\hat{w}, \frac{3}{2})$.

		After enough iterations of gradient descent on $f_S, f_{S'}$, we will obtain models $w_S$ and $w_{S'}$ that are close to the distinct local minima in the right-most graph in Figure \ref{quad_loss}. This will hold as long as $\gamma$ is not extremely large, in which case the steps of gradient descent could simply jump from one local minima to another. To ensure this does not happen, we restrict to $\gamma \leq 1$.

		Let $w_1 < w_2$ denote the two local minima of $g(w)$. For $z^* = (-1/2,1)$, plugging these values in to $\ell(w;z^*)$ shows that $|\ell(w_1;z^*)-\ell(w_2;z^*)| > 1$. Since $w_S$ is close to $w_1$ and $w_{S'}$ is close to $w_2$, we get the following theorem.
		\begin{theorem}\label{unstable_gd}
		For all $n$, for all step-sizes $0 < \gamma \leq 1$, there is a $K$ such that for all $k \geq K$, there are data sets $S, S'$ of size $n$ differing in one entry and a non-zero measure set of initial starting points such that if we perform $k$ iterations of gradient descent with step-size $\gamma$ on $S$ and $S'$ to get outputs $\mathcal{A}(S), \mathcal{A}(S')$ then there is a $z^*$ such that
		$$|\ell(\mathcal{A}(S);z^*)-\ell(\mathcal{A}(S');z^*)| \geq \frac{1}{2}.$$
		\end{theorem}

		Theorem \ref{unstable_gd} establishes that there exist simple non-convex settings, for which the uniform stability of gradient descent does not decrease with $n$. In light of the work in \cite{hardt2}, where the authors show that for very conservative step-sizes, SGD is stable on non-convex loss function, we might wonder whether SGD is stable in this setting with moderate step-sizes. We know that gradient descent is not stable, by Theorem \ref{unstable_gd}, for $\gamma = 1$. For simplicity of analysis, we focus on the case where $\gamma = 1$.

		Suppose we run SGD on the above $f_S, f_{S'}$ with step-size $1$ and initialize near $\hat{w}$ (we will be more concrete later about where we initialize). With probability $\frac{n-1}{n}$, the first iteration of SGD will use the same example for both $S$ and $S'$, either $z = (-1,1)$ or $z =(-1/2,1)$. Computing derivatives at $\hat{w}$ shows
		$$\dfrac{d}{dw}\ell(w; (-1,1))\bigg|_{w = \hat{w}} \approx -0.486254.$$
		$$\dfrac{d}{dw}\ell(w; (-\frac{1}{2},1))\bigg|_{w = \hat{w}} \approx 0.486254.$$
		In both cases, the slope is at least $0.4$. Therefore, there is some $\eta$ such that for all $w \in [\hat{w}-\eta, \hat{w}+\eta]$, 
		$$\dfrac{d}{dw}\ell(w; (-1,1)) < -0.4.$$
		$$\dfrac{d}{dw}\ell(w; (-\frac{1}{2},1)) > 0.4.$$
		Since we are taking $\gamma = 1$ and $\hat{w} \approx 0.598004$, this implies that with probability $\frac{n-1}{n}$, after one step of SGD we move outside of the interval $(0.5,1)$. This is important since this is the interval where $\ell(w;(-1,1))$ and $\ell(w;(-1/2,1))$ have slopes that point them towards distinct basins. Similarly, outside of this interval we also have that the slopes of $\ell(w; (z_{\pm},0))$ point towards the same basin. 

		Therefore, continuing to run SGD in this setting, even if we now decrease the step-size, will eventually lead us to the same basins of $f_S(w), f_{S'}(w)$. Let $w_S, w_{S'}$ denote the outputs of SGD in this setting after enough steps so that we get convergence to within $\frac{1}{n}$ of a local minima. If our first sample $z$ was $(-1,1)$, we will end up in the right basin, while if our first sample $z$ was $(-1/2,1)$, we will end up in the left basin. In particular, for $\epsilon$ small, $z_{\pm}$ are close enough that the minima of $f_S, f_{S'}$ are within $\frac{1}{n}$ of each other. Note that the minima $w_1, w_2$ that $w_S, w_{S'}$ are converging to are different. However, because they are in the same basin we know that for $\epsilon$ small, $z_{\pm}$ are close enough that $\|w_S-w_{S'}\| \leq O(\frac{1}{n})$. Therefore, we have
		$$\|w_S-w_{S'}\| \leq O\left(\frac{1}{n}\right).$$
		Note that in our proof of the instability of gradient descent, we only needed to look at $z$ satisfying $\|z\| \leq 2$ to see the instability. However, for SGD if we restrict to $\|z\| \leq 2$, then by compactness we know that $\ell(w;z)$ will be Lipschitz. Therefore, with probability $\frac{n-1}{n}$,
		$$\|\ell(w_S;z) - \ell(w_{S'};z)\| \leq L\|w_S-w_{S'}\| \leq O\left(\frac{1}{n}\right).$$

		With probability $\frac{1}{n}$, SGD first sees the example on which $S, S'$ differ. In this case, $w_S, w_{S'}$ may end up in different basins of the right-most graph in Figure \ref{quad_loss}. Restricting to $\|z\|\leq 2$, by compactness we have $|\ell(w_S;z) - \ell(w_{S'};z)| \leq C$ for some constant $C$. Therefore, $|\ell(w_S;z) - \ell(w_{S'}; z)| = O(1/n)$ with probability $1-1/n$ and $|\ell(w_S;z) - \ell(w_{S'};z)| = O(1)$ with probability $1/n$. This implies  the following theorem.

		\begin{theorem}Suppose that we initialize SGD in $[\hat{w}-\eta,\hat{w}+\eta]$ with a step-size of $\gamma = 1$. Let $\mathcal{A}(S), \mathcal{A}(S')$ denote the output of SGD after $k$ iteration for sufficiently large $k$. For $\|z\| \leq 2$, 
		$$\expect_{\mathcal{A}}\big[\ell(\mathcal{A}(S);z) - \ell(\mathcal{A}(S');z)\big] \leq O\left(\frac{1}{n}\right).$$\end{theorem}

		This is in stark contrast to gradient descent, which is unstable in this setting. While work in \cite{zhang2016understanding} suggests that this stability of SGD even in non-convex settings is a more general phenomenon, proving that this holds remains an open problem.

\section{Conclusion}\label{section_conc}

	The success of machine learning algorithms in practice is often dictated by their ability to generalize. While recent work has developed great insight in to the training error of machine learning algorithms, much less is understood about their generalization error. Most work up to this point has either focused on specific algorithms or has made assumptions on the loss function (such as strong convexity) that may not be true in practice. By analyzing stability as coming from the convergence of an algorithm to global minima and the geometry surrounding them, we are able to derive much broader results. We develop easy-to-use stability results that encompass a general class of non-convex settings, and some of those can appear in interesting learning setups. Our bounds establish the	stability for SGD, GD, SVRG, and RCD that is quantitatively comparable to more specialized results made in the past. 
	Although our bounds are not nuanced enough to explain the generalization of mini-batch SGD compared to large-batch SGD, or full-batch GD, we hope that the generality of our nonconvex bounds serves as a step towards developing stability analyses that may help in understanding the generalization performance of practical machine learning algorithms.

	There are still many exciting open problems concerning the stability and generalization of machine learning and optimization algorithms. We give a few below.

	\bigskip

	{\bf Stability for non-convex loss functions:} While our results establish the stability of learning algorithms in some non-convex scenarios, it is unclear how to extend  them directly to more general non-convex loss functions. Due to the wide variety of similar but not identical algorithms in machine learning, it would be particularly interesting to derive black-box results on the stability of learning algorithms for general non-convex loss functions, even when it concerns convergence to approximate local minima. 
	In general, we expect these to be a function of the geometry of the loss function and the convergence of the algorithm, but it is quite possible that there are other factors that more directly control stability.

	\bigskip

	{\bf Generalization error of local minima:} In general, we cannot expect a loss function to have only global minima. The question remains, among all local minima of a loss function, which one has the smallest generalization gap? The local minima with the smallest generalization error may not be a global minimizer. Even if we restrict to global minimizers, these may have different generalization errors. Is there a simple geometric characterization of their generalization error? This question may be connected to how {\it sharp} the loss function looks nearby the point. Steeper loss functions imply that small perturbations can greatly change the error, which suggests smaller generalization error. Theoretical results concerning this fact could be useful in stability analysis and the design of algorithms.

	\bigskip

	{\bf Generalization of SGD vs GD:} We showed above that there are settings in which gradient descent is not uniformly stable, but SGD is. Empirically, SGD leads to small generalization error in neural networks \cite{zhang2016understanding}. Theoretically, it is unclear how widespread this phenomenon is. Does SGD actually lead to more generalizable models than gradient descent? If so, why? How does this compare to other variants of SGD? Any theoretical results concerning the difference in generalization between SGD and other related algorithms could be extremely useful in the design of training algorithms and for understanding the empirical success of neural networks trained by SGD.

	\bigskip

	{\bf The geometry of critical points in real neural networks:} We show above that linear neural networks obey relatively nice conditions on their local minima. In particular, as long as we restrict to weights that are full rank, all critical points are actually global minima. However, linear neural networks are not very exciting (\eg they just correspond to linear models). 
Developing general theorems concerning the geometry underlying the critical points of neural networks is a challenging but very interesting open problem.

\bibliographystyle{plain} 
\bibliography{stab}

\newpage

\appendix

\section{Omitted Proofs}

\subsection{Properties of PL and QG Functions}

	We have the following equivalent definition of PL due to Karimi et al.

	\begin{lemma}[Error bound, \cite{karimi}]
	\label{equiv-pl}The PL condition is equivalent to the condition that there is some constant $\mu > 0$ such that for all $x$, $\|\nabla f(x)\| \geq \mu\|x_p-x\|$.\end{lemma}		

	In this same paper, Karimi et al. show that PL functions also satisfy QG.

	\begin{lemma}[\cite{karimi}]\label{pl-implies-qg}The PL condition implies the QG condition.\end{lemma}

\subsection{Proof of Theorem \ref{thm:black-box-pl-ptwise}}\label{proof:bb-pl-ptwise}

\begin{proof}
			Fix a training set $S$ and $i \in \{1,\ldots, n\}$. We will show pointwise hypothesis stability for all $S, i$ instead of for them in expectation. Let $w_1$ denote the output of $\mathcal{A}$ on $S$, and let $w_2$ denote the output of $\mathcal{A}$ on $S^{i}$. Let $w_1^*$ denote the critical point of $f_S$ to which $w_1$ is approaching, and $w_2^*$ denote the critical point of $f_{S^i}$ that $w_2$ is approaching. We then have,
			\begin{align*}
			&|\ell(w_1;z_i)-\ell(w_2;z_i)|\\
			&\leq |\ell(w_1;z_i)-\ell(w_1^*;z_i)| + |\ell(w_1^*;z_i)-\ell(w_2^*;z_i)| + |\ell(w_2^*;z_i)-\ell(w_2;z_i)|\numberthis\label{three_sum_local}
			\end{align*}

			We first wish to bound the first and third terms of (\ref{three_sum_local}). The bound depends on the case in Theorem \ref{thm:black-box-pl-ptwise}.

			{\bf Case 1:} By assumption, $\|w_1-w_1^*\| = O(\epsilon_{\mathcal{A}})$. Since $\ell(\cdot;z_i)$ is $L$-Lipschitz, this implies
			$$|\ell(w_1;z_i)-\ell(w_1^*;z_i)| \leq L\|w_1-w_1^*\| = O(L\epsilon_{\mathcal{A}}).$$

			{\bf Case 2:} As stated in Lemma \ref{pl-implies-qg}, the PL condition implies the QG condition. Therefore,
			\begin{equation}\frac{\mu}{2}\|w_1-w_1^*\|^2 \leq |f_S(w_1)-f_S(w_1^*)|.\end{equation}

			By assumption on case 2, $|f_S(w_1)-f_S(w_1^*)| \leq O(\epsilon'_{\mathcal{A}})$. This implies
			\begin{align*}
			\|w_1-w_1^*\| &\leq \frac{\sqrt{2}}{\sqrt{\mu}}\sqrt{|f_S(w_1)-f_S(w_1^*)|}\\
			&= O\bigg(\sqrt{\frac{\epsilon'_{\mathcal{A}}}{\mu}}\bigg)\end{align*}

			{\bf Case 3:} By Lemma \ref{equiv-pl}, the PL condition on $w_1, w_1^*$ implies that
			$$\|\nabla f_S(w_1) \| \geq \mu\|w_1-w_1^*\|.$$

			Using the fact that $f_S$ is $L$-Lipschitz and the fact that $\|\nabla f_S(w_j)\| \leq O(\epsilon''_{\mathcal{A}})$ by assumption on Case 3, we find

			$$|\ell(w_1;z_i)-\ell(w_1^*;z_i)| \leq L\|w_1-w_1^*\| \leq \dfrac{L}{\mu}\|\nabla f_S(w_1)\| \leq O\bigg(\frac{L\epsilon''_{\mathcal{A}}}{\mu}\bigg).$$

			In the above three cases, we can bound $|\ell(w_2;z_i)-\ell(w_2^*;z_i)|$ in the same manner. We now wish to bound the second term of (\ref{three_sum_local}). Note that we can manipulate this term as

			\begin{align*}
			|\ell(w_1^*;z_i)-\ell(w_2^*;z_i)| &= |(nf_S(w_1^*)-(n-1)f_{S^i}(w_1^*))-(nf_S(w_2^*)+(n-1)f_{S^i}(w_2^*))|\\
			&\leq n|f_S(w_1^*)-f_S(w_2^*)| + (n-1)|f_{S^i}(w_1^*)-f_{S^i}(w_2^*)|.\numberthis \label{middle_term}
			\end{align*}

			By the PL condition, we can find a local minima $u$ of $f_S$ such that
			$$\|\nabla f_S(w_2^*)\|^2 \geq \mu|f_S(w_2^*)-f_S(u)|.$$

			Similarly, we can find a local minima $v$ of $f_{S^i}$ such that
			$$\|\nabla f_{S^i}(w_1^*)\|^2 \geq \mu|f_{S^i}(w_1^*)-f_{S^i}(v)|.$$

			Note that since $\nabla f_{S^i}(w_2^*) = 0$, we get:
			$$\|\nabla f_S(w_2^*)\|^2 = \frac{1}{n^2}\|\nabla \ell(w_2^*;z_i)\|^2 \leq \frac{L^2}{n^2}.$$

			Similarly, since $\nabla f_S(w_1^*) = 0$, we get:
			$$\|\nabla f_{S^i}(w_1^*)\|^2 = \frac{1}{(n-1)^2}\|\nabla \ell(w_1^*;z_i)\|^2 \leq \frac{L^2}{(n-1)^2}.$$

			Since all local minima of a PL function are global minima, we obtain
			\begin{align*}
			n|f_S(w_1^*)-f_S(w_2^*)| &\leq n|f_S(w_1^*)-f_S(u)| + n|f_S(u)-f_S(w_2^*)|\\
			&\leq  n\frac{L^2}{\mu n^2}\\
			&= \frac{L^2}{\mu n}.\numberthis \label{middle_term_1}\end{align*}

			In a similar manner, we get
			\begin{align*}
			(n-1)|f_{S^i}(w_1^*)-f_{S^i}(w_2^*)| &\leq (n-1)|f_{S^i}(w_1^*)-f_{S^i}(v)| + (n-1)|f_{S^i}(v)-f_{S^i}(w_2^*)|\\
			&\leq (n-1)\frac{L^2}{\mu(n-1)^2}\\
			&\leq \frac{L^2}{\mu (n-1)}.\numberthis \label{middle_term_2}\end{align*}

			Plugging in \ref{middle_term_1}, \ref{middle_term_2} in to \ref{middle_term}, we find
			\begin{align*}
			|\ell(w_1^*;z_i)-\ell(w_2^*;z_i)| &\leq \dfrac{L^2}{\mu n} + \dfrac{L^2}{\mu(n-1)}.\end{align*}

			This proves the desired result.

		\end{proof}

\subsection{Proof of Theorem \ref{thm:black-box-qg-ptwise}}

\begin{proof}

			Fix a training set $S$ and $i \in \{1,\ldots, n\}$. We will show pointwise hypothesis stability for all $S, i$ instead of for them in expectation. Let $w_1$ denote the output of $\mathcal{A}$ on $S$, and let $w_2$ denote the output of $\mathcal{A}$ on $S^{i}$. Let $w_1^*$ denote the critical point of $f_S$ to which $w_1$ is approaching, and $w_2^*$ denote the critical point of $f_{S^i}$ that $w_2$ is approaching. We then have,
			\begin{align*}
			&|\ell(w_1;z_i)-\ell(w_2;z_i)|\\
			&\leq |\ell(w_1;z_i)-\ell(w_1^*;z_i)| + |\ell(w_1^*;z_i)-\ell(w_2^*;z_i)| + |\ell(w_2^*;z_i)-\ell(w_2;z_i)|\numberthis\label{three_sum_local_qg}
			\end{align*}

			We first wish to bound the first and third terms of (\ref{three_sum_local_qg}). The bound depends on the case in Theorem \ref{thm:black-box-qg-ptwise}.

			{\bf Case 1:} By assumption, $\|w_1-w_1^*\| = O(\epsilon_{\mathcal{A}})$. Since $\ell(\cdot;z_i)$ is $L$-Lipschitz, this implies
			$$|\ell(w_1;z_i)-\ell(w_1^*;z_i)| \leq L\|w_1-w_1^*\| = O(L\epsilon_{\mathcal{A}}).$$

			{\bf Case 2:} By the QG condition, we have
			\begin{equation}\frac{\mu}{2}\|w_1-w_1^*\|^2 \leq |f_S(w_1)-f_S(w_1^*)|.\end{equation}

			By assumption on case 2, $|f_S(w_1)-f_S(w_1^*)| \leq O(\epsilon'_{\mathcal{A}})$. This implies
			\begin{align*}
			\|w_1-w_1^*\| &\leq \frac{\sqrt{2}}{\sqrt{\mu}}\sqrt{|f_S(w_1)-f_S(w_1^*)|}\\
			&= O\bigg(\sqrt{\frac{\epsilon'_{\mathcal{A}}}{\mu}}\bigg)\end{align*}

			In the above two cases, we can bound $|\ell(w_2;z_i)-\ell(w_2^*;z_i)|$ in the same manner. We now wish to bound the second term of (\ref{three_sum_local_qg}). By the QG property, we can pick some local minima $v$ of $f_S$ such that
			\begin{equation}\label{qg_pt_1}
			||w_2^*-v|| \leq \frac{2}{\sqrt{\mu}}\sqrt{|f_S(w_2^*)-f_S(v)|}.\end{equation}

			We then have
			\begin{align*}
			|\ell(w_1^*;z_i)-\ell(w_2^*;z_i)| &\leq |\ell(w_1^*;z_i) - \ell(v;z_i)| + |\ell(v;z_i)-\ell(w_2^*;z_i)|.\end{align*}

			Note that by assumption, $f_S(w_1^*) = f_S(u) = 0$, so $|\ell(w_1^*;z_i) - \ell(v;z_i)| = 0$. By the Lipschitz property and the QG condition we have
			\begin{align*}
			|\ell(v;z_i)-\ell(w_2^*;z_i)| &\leq L||v-w_2^*||\\
			&\leq \dfrac{2L}{\sqrt{\mu}}\sqrt{|f_S(w_2^*)-f_S(v)|}\\
			&\leq \dfrac{2L}{\sqrt{\mu}}\sqrt{|f_S(w_2^*)-f_S(w_1^*)| + |f_S(w_1^*)-f_S(v)|}.\numberthis\label{sqrt_bound}\end{align*}

			Note that $|f_S(w_1^*)-f_S(v)| = 0$ by our realizability assumption. By assumption on $w_1^*,w_2^*$, we know that $f_S(w_1^*) \leq f_S(w_2^*)$ and $f_{S^i}(w_2^*) \leq f_{S^i}(w_1^*)$. Some simple analysis shows
			\begin{align*}
			nf_S(w_2^*) &= nf_{S^i}(w_2^*) + \ell(w_2^*;z_i)\\
			&\leq nf_{S^i}(w_1^*) + \ell(w_2^*;z_i)\\
			&= nf_S(w_1^*) +\ell(w_2^*;z_i)-\ell(w_1^*;z_i).\end{align*}

			Since $\ell(w_2^*;z_i) \leq c$, this implies that $n|f_S(w_2^*)-f_S(w_1^*)| \leq c$. Plugging this bound in to \ref{sqrt_bound}, we get
			$$|\ell(v;z_i)-\ell(w_2^*;z_i)| \leq 2L\sqrt{\frac{c}{\mu n}}.$$

			This proves the desired result.
		\end{proof}

\subsection{Proof of Theorem \ref{thm:black-box-pl}}

	Let $S_1 = \{z_1,\ldots, z_n\}, S_2 = \{z_1,\ldots, z_{n-1},z_n'\}$ be data sets of size $n$ differing only in one entry. Let $w_i$ denote the output of $\mathcal{A}$ on data set $S_i$ and let $w_i^*$ denote $w_{S_i}^*$. Let $f_i(w) = f_{S_i}(w)$.

\begin{proof}
		Using the fact that $\ell(\cdot;z)$ is $L$-Lipschitz we get
		\begin{align*}
		|\ell(w_1;z) - \ell(w_2;z)| &\leq L\|w_1-w_2\|\\
		&\leq L\|w_1-w_1^*\| + L\|w_1^*-w_2^*\| + L\|w_2^*-w_2\|\numberthis\label{three_sum}\end{align*}

		Note that by {\bf A1}, we know that $w_1^*$ is the closest optimal point of $f_1$ to $w_2^*$. By the PL condition,
		\begin{align*}
		\|w_1^*-w_2^*\| & \leq \frac{1}{\mu}\|\nabla f_1(w_2^*)\|\\
		&= \frac{1}{\mu}\|\nabla f_2(w_2^*)-\frac{1}{n}\nabla \ell(w_2^*;z_n') + \frac{1}{n}\nabla \ell(w_2^*;z_n)\|\\
		&\leq \frac{1}{\mu n}(\|\nabla \ell(w_2^*;z_n')\| + \|\nabla \ell(w_2^*;z_n)\|)\\
		&\leq \frac{2L}{\mu n}\end{align*}

		This bounds the second term of \ref{three_sum}. The first and third terms must be bounded differently depending on the case.

		{\bf Case 1:} By assumption, for $i = 1,2$, $\|w_i-w_i^*\| = O(\epsilon_{\mathcal{A}})$, proving the result.

		{\bf Case 2:} As stated in Lemma \ref{pl-implies-qg}, PL implies QG. Therefore for $i = 1,2$,
		\begin{align*}
		\|w_i-w_i^*\| &\leq \frac{\sqrt{2}}{\sqrt{\mu}}\sqrt{f_i(w_i)-f_i^*}\\
		&= O\bigg(\sqrt{\frac{\epsilon'_{\mathcal{A}}}{\mu}}\bigg)\end{align*}

		{\bf Case 3:} As mentioned above in Lemma \ref{equiv-pl}, the PL condition implies that for $i = 1,2$,
		$$\|\nabla f_i(w_i) \| \geq \mu\|w_i-w_i^*\|.$$

		Since $\|\nabla f_i(w_i)\| \leq O(\epsilon''_{\mathcal{A}})$, we get the desired result.\end{proof}

\subsection{Proof of Theorem \ref{thm:black-box-qg}}

	We will use the following lemma.

	\begin{lemma}\label{lem:qg-bound}
		Let $f_S$ be QG and assume that  $\ell(w;z) \leq c$ for all $z$ and $w \in \mathcal{X}$. We assume {\bf A1} as above. Then for $S_1, S_2$ differing in at most one place,

		$$\|w_1^*-w_2^*\| \leq 2\sqrt{\frac{c}{\mu n}}.$$
		\end{lemma}

		\begin{proof}
		By QG:
		\begin{align*}
		\frac{\mu}{2}\|w_1^*-w_2^*\|^2 &\leq |f_1(w_2^*)-f_1(w_1^*)|\\
		&\leq |f_1(w_2^*)-f_2(w_2^*)| + |f_2(w_2^*)-f_1(w_1^*)|\end{align*}

		Note that for all $w$, $|f_1(w)-f_2(w)| = \frac{1}{n}|\ell(w;z_n)-\ell(w;z_n')|$, so this is bounded by $\frac{c}{n}$.

		By this same reasoning we get:
		\begin{align*}
		f_2(w_2^*) \leq f_2(w_1^*) \leq f_1(w_1^*) + \frac{c}{n}\end{align*}

		The desired result follows.\end{proof}

		\begin{proof}[Proof of Theorem \ref{thm:black-box-qg}] Using the fact that $\ell(\cdot;z)$ is $L$-Lipschitz we get
		\begin{align*}
		|\ell(w_1;z) - \ell(w_2;z)| &\leq L\|w_1-w_2\|\\
		&\leq L\|w_1-w_1^*\| + L\|w_1^*-w_2^*\| + L\|w_2^*-w_2\|\numberthis\label{three_sum_qg}\end{align*}

		Note that by {\bf A1}, we know that $w_1^*$ is the closest optimal point of $f_1$ to $w_2^*$. By Lemma \ref{lem:qg-bound},
		\begin{align*}
		\|w_1^*-w_2^*\| & \leq 2\sqrt{\frac{c}{\mu n}}.\end{align*}

		This bounds the second term of \ref{three_sum}. The first and third terms must be bounded differently depending on the case.

		{\bf Case 1:} By assumption, for $i = 1,2$, $\|w_i-w_i^*\| = O(\epsilon_{\mathcal{A}})$, proving the result.

		{\bf Case 2:} By the QG property, we get that for $i = 1, 2$,
		\begin{align*}
		\|w_i-w_i^*\| &\leq \frac{\sqrt{2}}{\sqrt{\mu}}\sqrt{f_i(w_i)-f_i^*}\\
		&= O\bigg(\sqrt{\frac{\epsilon'_{\mathcal{A}}}{\mu}}\bigg)\end{align*}

		This proves the desired bound.\end{proof}

\subsection{Stability of Gradient Descent for Convex Loss Functions}\label{stab_gd_section}
\label{sec:gd_stab}

		To prove the stability of gradient descent, we will assume that the underlying loss function is smooth.
		\begin{defn}A function $f: \Omega \to \real$ is $\beta$-smooth if for all $u, v \in \Omega$, we have
		$$\|\nabla f(u)-\nabla f(v)\| \leq \beta\|u-v\|.$$\end{defn}

		In \cite{hardt2}, Hardt et al. show the following theorem.
		\begin{theorem}[\cite{hardt2}]\label{stab_sgd}
		Let $\ell(\cdot;z)$ be $L$-Lipschitz, $\beta$-smooth, and convex for all $z$. Say we perform $T$ iterations of SGD with a constant step size $\gamma \leq \nicefrac{2}{\beta}$ to train iterates $w_t$ on $S$ and $\hat{w}_t$ on $S'$. Then for all such $S, S'$ with $|S| = |S'| = n$ such that $S, S'$ differ in at most one example,
		$$\expect_{\mathcal{A}}[\|w_{T}-\hat{w}_T\|] \leq \frac{2\gamma LT}{n}$$

		If $\ell(\cdot;z)$ is $\lambda$-strongly convex for all $z$, then
		$$\expect_{\mathcal{A}}[\|w_T-\hat{w}_T\|] \leq \frac{2L}{\lambda n}$$\end{theorem}

		Performing similar analysis for gradient descent, we obtain the following theorem.

		\begin{theorem}\label{stab_gd}
		Assume that for all $z$, $\ell(\cdot ; z)$ is convex, $\beta$-smooth, and $L$-Lipschitz. Say we run GD for $T$ iterations with step sizes $\gamma_t$ such that $\gamma_t \leq \frac{2}{\beta}$. Then GD is uniformly stable with
		\begin{align*}
		\epsilon_{\text{stab}} \leq \dfrac{2L^2}{n}{\sum_{t=0}^T \gamma_t}.\end{align*}
		If $\ell(\cdot;z)$ is $\lambda$-strongly convex for all $z$, then GD is uniformly stable with
		$$\epsilon_{stab} \leq \frac{2L}{\lambda n}.$$
		\end{theorem}			

		To prove this theorem, we use similar techniques to those in \cite{hardt2}. We first consider the convex case.

		\begin{proof}
			By direct computation, we have:
			\begin{align*}
			 \|w_T-\hat{w}_{T}\| &= \|w_{T-1}-\hat{w}_{T-1} - \frac{\gamma_T}{n}\sum_{i=1}^{n-1} \bigg(\nabla\ell(w_{T-1} ; z_i) - \grad\ell(\hat{w}_{T-1};z_i)\bigg)\\&\hspace{10pt}  - \frac{\gamma_T}{n}\nabla \ell(w_{T-1} ; z_n) + \frac{\gamma_T}{n}\nabla \ell(\hat{w}_{T-1} ; z_n')\|\\
			& \leq \|w_{T-1}-\hat{w}_{T-1} - \frac{\gamma_T}{n}\sum_{i=1}^{n-1} \bigg(\nabla \ell(w_{T-1} ; z_i) - \grad\ell(\hat{w}_{T-1};z_i)\bigg)\|\\&\hspace{10pt}+ \|\frac{\gamma_T}{n}\nabla \ell(w_{T-1} ; z_n) - \frac{\gamma_T}{n}\nabla \ell(\hat{w}_{T-1} ; z_n')\|.
			\end{align*}

			Note that since $\ell(\cdot;z)$ is $L$-Lipschitz, the second part of this summand is bounded by $\displaystyle\frac{2\gamma_T L}{n}$. We now wish to bound the first part. We get:

			\begin{align*}
			& \|w_{T-1}-\hat{w}_{T-1} - \frac{\gamma_T}{n}\sum_{i=1}^{n-1} \bigg(\grad \ell(w_{T-1} ; z_i) - \grad\ell(\hat{w}_{T-1};z_i)\bigg)\|^2\\
			&= \|w_{T-1}-\hat{w}_{T-1}\|^2 - \frac{2\gamma_T}{n}\langle w_{T-1}-\hat{w}_{T-1}, \sum_{i=1}^{n-1} \bigg(\grad \ell(w_{T-1} ; z_i) - \grad\ell(\hat{w}_{T-1};z_i)\bigg)\rangle\\
			& \hspace{2cm}+\frac{\gamma_T^2}{n^2}\|\sum_{i=1}^{n-1} \bigg(\grad \ell(w_{T-1} ; z_i) - \grad\ell(\hat{w}_{T-1};z_i)\bigg)\|^2\\
			&\leq \|w_{T-1} - \hat{w}_{T-1}\|^2- \sum_{i=1}^{n-1} \bigg(\frac{2\gamma_T}{n^2}\langle w_{T-1} - \hat{w}_{T-1}, \grad \ell(w_{T-1} ; z_i) - \grad\ell(\hat{w}_{T-1};z_i)\rangle\bigg)\\
			&\hspace{2cm} +\sum_{i=1}^{n-1}\bigg(\frac{\gamma_T^2}{n^2}\|\grad\ell(w_{T-1} ; z_i) - \grad\ell(\hat{w}_{T-1};z_i)\|^2\bigg)\\
			&\leq \|w_{T-1} - \hat{w}_{T-1}\|^2 +\sum_{i=1}^{n-1}(\frac{\gamma_T^2}{n^2}-\frac{2\gamma_T}{n^2\beta})\|\grad\ell(w_{T-1} ; z_i) - \grad\ell(\hat{w}_{T-1};z_i)\|^2.
			\end{align*}

			Note that this last step holds by co-coercivity of the gradient of $\ell(\cdot, z_i)$. In particular, if $\gamma_T \leq \frac{2}{\beta}$, each of the $n-1$ summands on the right will be nonpositive. Therefore for such $\gamma_T$, we get:
			\begin{align*}
			\|w_{T-1}-\hat{w}_{T-1} - \frac{\gamma_T}{n}\sum_{i=1}^{n-1} \bigg(\grad \ell(w_{T-1} ; z_i) - \grad\ell(\hat{w}_{T-1};z_i)\bigg)\| \leq \|w_{T-1}-\hat{w}_{T-1}\|.\end{align*}

			So, if $\frac{\gamma_t}{n} \leq \frac{2}{\beta}$ for all $t$, we get:
			\begin{align*}
			\|w_T-\hat{w}_{T}\| & \leq \|w_{T-1}-\hat{w}_{T-1}\| + \frac{2\gamma_TL}{n}\\
			& \leq \|w_{T-2} - \hat{w}_{T-2}\| + \frac{2L}{n}(\gamma_{T-1}+\gamma_T)\\
			& \vdots\\
			& \leq \|w_0-\hat{w}_0\| + \frac{2L}{n}\sum_{t=1}^T\gamma_t\\
			&=\frac{2L}{n}\sum_{t=1}^T\gamma_t.
			\end{align*}

			This last step follows from the fact that $w_0 = \hat{w}_0$ if we initialize at the same point. Using the fact that $f(w)$ is $L$-Lipschitz (since $\ell$ is), we get:
			\begin{align*}
			|f(w_T ; z) - f(\hat{w}_T ; z)| \leq \frac{2L^2}{n}\sum_{t=0}^n\gamma_t.\end{align*}
		\end{proof}	

		We now move to the $\lambda$-strongly convex case. For simplicity of analysis, we assume that we use a constant step size $\gamma$ such that $\gamma \leq 1/\beta$.

		\begin{proof}
			The proof remains the same, except when using co-coercivity. Under this assumption, some plug and play in an analogous fashion will show:
			\begin{align*}
			& \|w_{T-1}-\hat{w}_{T-1} - \frac{\gamma_T}{n}\sum_{i=1}^{n-1} \bigg(\grad \ell(w_{T-1} ; z_i) - \grad\ell(\hat{w}_{T-1};z_i)\bigg)\|^2\\
			& \leq \bigg(1 - \frac{2\gamma\lambda\beta}{\lambda+\beta}\bigg)\|w_{T-1}-\hat{w}_{T-1}\|^2\\
			&\hspace{2cm} +\sum_{i=1}^{n-1}(\frac{\gamma_T^2}{n^2}-\frac{2\gamma_T}{n\beta})\|\grad\ell(w_{T-1} ; z_i) - \grad\ell(\hat{w}_{T-1};z_i)\|^2.
			\end{align*}

			Note that if $\gamma \leq \frac{1}{\beta}$ then the second term is nonnegative and one can show that this implies:
			\begin{align*}
			& \|w_{T-1}-\hat{w}_{T-1} - \frac{\gamma_T}{n}\sum_{i=1}^{n-1} \bigg(\grad \ell(w_{T-1} ; z_i) - \grad\ell(\hat{w}_{T-1};z_i)\bigg)\|\\
			& \leq \bigg(1-\gamma\lambda\bigg)\|w_{T-1}-\hat{w}_{T-1}\|.\end{align*}

			Combining, this shows:

			\begin{align*}
			\|w_{T}-\hat{w}_{T-1}\| & \leq (1-\gamma\lambda)\|w_{T-1}-\hat{w}_{T-1}\| + \frac{2L\gamma}{n}\\
			&\leq (1-\gamma\lambda)^2\|w_{T-2}-\hat{w}_{T-2}\| + \frac{2L\gamma}{n}\bigg(1 + (1-\gamma\lambda)\bigg)\\
			& \hspace{2cm}\vdots\\
			& \leq \frac{2L\gamma}{n}\sum_{t=0}^T(1-\gamma\lambda)^t\\
			& \leq \frac{2L}{\lambda n}.\end{align*}

		This implies uniform stability with parameter $\frac{2L^2}{\lambda n}$.\end{proof}

\subsection{Proof of Theorem \ref{thm:str-cvx-ReLU}}\label{proof:str-cvx-relu}

		\begin{proof}
			For almost all $w$, we can write $\sigma(Xw)$ as $\diag(b)Xw$ for a vector $b$ where $b_i(Xw)_i = \sigma((Xw)_i)$ (this only exludes points $w$ such that $(Xw)_i$ is on a cusp of the piecewise-linear function). Then in an open neighborhood of such an $w$, we find:
			$$f(w) = g(\sigma(Xw)) = g(\diag(b)Xw)$$

			For a given $w$, let $w_p$ be the closest global minima of $f$ (\ie the closest point such that $f^* = f(w_p)$). By strong convexity of $g$, we find:
			\begin{align*}
			& g(\diag(b)Xw_p) \geq g(\diag(b)Xw) + \langle \nabla g(\diag(b)Xw), \diag(b)X(w_p-w)\rangle + \frac{\lambda}{2}\|\diag(b)X(w_p-w)\|^2\\
			\implies & g(\diag(b)Xw_p) \geq g(\diag(b)Xw) +\langle X^T\diag(b)^T\nabla g(\diag(b)Xw), w_p-w\rangle + \frac{\lambda}{2}\|\diag(b)X(w_p-w)\|^2\\
			\implies & f(w_p) \geq f(w) + \langle \nabla f(w), w_p-w\rangle + \frac{\lambda \sigma_{\min}(X)^2\sigma_{\min}(\diag(b))^2}{2}\|w_p-w\|^2\end{align*}

			Note that the minimum singular value of $\diag(b)$ is the square root of the minimum eigenvalue of $\diag(b)^2$. Since $\diag(b)^2$ has entries $c_i^2$ on the diagonal, we know that the minimum singular value is at least $c = \min_i\{|c_i|\}$. Therefore we get:
			\begin{align*}
			f(w_p) &\geq f(w) + \langle f(w), w_p-w\rangle +\frac{\lambda\sigma_{\min}(X)^2c^2}{2}\|w_p-w\|^2\\
			&\geq f(w) + \min_y\bigg[\langle \nabla f(w),y-w\rangle + \dfrac{\lambda\sigma_{\min}(X)^2c^2}{2}\|y-w\|^2\bigg]\\
			&= f(w) -\frac{1}{2\lambda\sigma_{\min}(X)^2c^2}\|\nabla f(w)\|^2.
			\end{align*}
		\end{proof}

\subsection{Proof of Lemma \ref{lem:proj-bound}}\label{proof:proj-bound}

	\begin{proof}
		Using basic properties of the Frobenius norm and the definition of the pseudo-inverse, we have
		\begin{align*}
		\|(WX-Y)X^T\|_{F}^2\|(XX^T)^{-1}X\|_{F}^2 & \geq \|(WX-Y)X^T(XX^T)^{-1}X\|_F^2\\
		& = \|(WX-Y)X^+X\|_{F}^2\\
		&= \|WXX^+X-YX^+X\|_{F}^2\\
		&= \|WX-YX^+X\|_F^2.
		\end{align*}
		This last step follows by basic properties of the pseudo-inverse. By the triangle inequality,
		$$\|WX-Y\|_F^2 = \|WX-YX^+X+YX^+X-Y\|_F^2 \leq \|YX^+X-Y\|_F^2 + \|WX-YX^+X\|_F^2.$$
		Note that $YX^+X-Y$ is the component of $Y$ that is orthogonal to the row-space of $X$, while $YX^+X$ is the projection of $Y$ on to this row space. Therefore, $YX^+X-Y$ is orthogonal to $WX-YX^+X$ with respect to the trace inner product. Therefore, the inequality above is actually an equality, that is
		$$\|WX-Y\|_F^2 = \|YX^+X-Y\|_F^2 + \|WX-YX^+X\|_F^2.$$
		Putting this all together, we find
		$$\|(XX^T)^{-1}X\|_{F}^2 \|(WX-Y)X^T\|_{F}^2 \geq \|WX-Y\|_F^2 - \|YX^+X-Y\|_F^2.$$
		\end{proof}

\subsection{Proof of Lemma \ref{lem:linear-grad-bound}}\label{proof:linear-grad-bound}

	\begin{proof}
			Our proof uses similar techniques to that of Hardt and Ma \cite{hardt1}. We wish to compute the gradient of $f$ with respect to a matrix $W_j$. One can show the following:
			$$\dfrac{\partial f}{\partial W_j} = W_{j+1}^T\ldots W_\ell^T(WX-Y)X^TW_1^T\ldots W_{j-1}^T.$$

			Using the fact that for a matrix $A \in \real^{d\times d}$ and another matrix $B \in \real^{d\times k}$, we have $\|AB\|_{F} \geq \sigma_{\min}(A)\|B\|_{F}$, we find:

			\begin{align*}
			\bigg|\bigg|\dfrac{\partial f}{\partial W_j}\bigg|\bigg|_{F} &\geq \prod_{i \neq j}\sigma_{\min}(W_i)\|(WX-Y)X^T\|_{F}.\end{align*}

			By assumption, $\sigma_{\min}(W_j) \geq \tau$. Therefore:
			\begin{align*}
			\bigg|\bigg|\dfrac{\partial f}{\partial W_j}\bigg|\bigg|_{F}^2 \geq \tau^{2\ell-2}\|(WX-Y)X^T\|_{F}^2.\end{align*}

			Taking the gradient with respect to all $W_i$ we get:
			\begin{align*}
			\bigg|\bigg|\dfrac{\partial f}{\partial (W_1,\ldots, W_\ell)}\bigg|\bigg|_{F}^2 &= \sum_{j=1}^\ell \bigg|\bigg|\dfrac{\partial f}{\partial W_j}\bigg|\bigg|_{F}^2\\
			& \geq \ell \tau^{2\ell-2}\|(WX-Y)X^T\|_{F}^2.\end{align*}
		\end{proof}

	\subsection{SGD with stepsizes $\gamma_t = c/t$ can be slow on smooth functions}
		\label{appendix:sgd_exp_time}
		In this subsection we show that a provable rate for SGD on smooth functions with learning rate proportional to $O(1/t)$ might require a large number of iterations. While \cite{hardt2} show stability of SGD in non-convex settings with such a step-size, their stability bounds grow close to linearly with the number of iterations. When exponentially many steps are taken, this no longer implies useful generalization bounds on SGD.

		When a function $f(x) = \sum_{i=1}^n f_i(x)$ is $\beta$ smooth on its domain, then the following holds:
		$$f(x)-f(y)\le \langle\nabla f(y), x-y\rangle + \frac{\beta}{2} \|x-y\|^2.$$ 
		Fix some $t$ and let $x = x_{t+1} = x_t -\gamma_t\nabla f_{s_{t}} (x_t)$ and let $y = x_t$. Here, $x_0$ is set to some initial vector value, $s_t$ is a uniform iid sample from $\{1,\ldots, n]$, and $\gamma_t = c/t$ for some constant $c>0$. 
		Then, due to the $\beta$-smoothness of $f$ we have the following:
		\begin{align*}
		f(x_{t+1})-f(x_t)
		&\le \langle\nabla f(x_t), x_{t+1}-x_t\rangle + \frac{\beta\gamma_t^2}{2} \|\nabla f_{s_t}(x_t)\|^2\\
		&\le -\gamma_t \langle\nabla f(x_t),\nabla f_{s_t}(x_t) \rangle + \frac{\beta\gamma_t^2}{2} \|\nabla f_{s_t}(x_t)\|^2.
		\end{align*}
		Taking expectation with respect to all random samples $s_t$, yields
		\begin{align*}
		&\mathbb{E}[f(x_{t+1})-f(x_t)] \le -\gamma_t \mathbb{E}[\|\nabla f(x_t)\|^2] + \frac{\beta\gamma_t^2}{2} \mathbb{E}[\|\nabla f(x_t)\|]\\
		\Leftrightarrow &
		\left(\gamma_t- \frac{\beta\gamma_t^2}{2} \right)\mathbb{E}[\|\nabla f(x_t)\|^2]  \le \mathbb{E}\left[f(x_t)-f(x_{t+1})\right].
		\end{align*}
		Summing the above inequality for all $t$ terms from 0 to $T$ we get
		\begin{align*}
		&\sum_{t=1}^T\left(\gamma_t- \frac{\beta\gamma_t^2}{2} \right)\mathbb{E}[\|\nabla f(x_t)\|^2]  \le \mathbb{E}[f(x_0)-f(x_T)]\\
		\Rightarrow &
		\sum_{t=1}^T\left(\gamma_t- \frac{\beta\gamma_t^2}{2} \right)\mathbb{E}[\|\nabla f(x_t)\|^2]  \le f(x_0)\\
		\Rightarrow &
		\min_{t=1,\ldots, T}\mathbb{E}[\|\nabla f(x_t)\|^2]  \le \frac{f(x_0)}{\sum_{t=1}^T\left(\gamma_t- \frac{\beta\gamma_t^2}{2} \right)}.
		\end{align*}
		Assuming that $\gamma_t=c/t$, we have
		\begin{align*}
		&\min_{t=1,\ldots, T}\mathbb{E}[\|\nabla f(x_t)\|^2]  \le \frac{f(x_0)}{\sum_{t=1}^Tc\left(\gamma_t- \frac{\beta c^2}{2t^2} \right)}\\
		\Rightarrow &
		\min_{t=1,\ldots, T}\mathbb{E}[\|\nabla f(x_t)\|^2]  \le \frac{f(x_0)}{c\sum_{t=1}^T t^{-1}}\\		
		\Rightarrow &
		\min_{t=1,\ldots, T}\mathbb{E}[\|\nabla f(x_t)\|^2]  \le \frac{f(x_0)}{C_1\log(T)},\\
		\end{align*}
		where $C_1$ is a universal constant that depends only on $c$. Observe that even if $C_2=0$, using the above simple bounding technique (a simplified version of the nonconvex convergence bounds of \cite{ghadimi2013stochastic}), requires $ O(e^{-\epsilon})$ steps to reach error $\epsilon$, \ie an exponentially large number of steps.  

		We note that the above bound does not imply that there does not exist a smooth function for which $1/t$ stepsizes suffice for polynomial-time convergence (in fact there are several convex problems for wich $1/t$ suffices for fast convergence). However, the above implies that when we are only assuming smoothness on a nonconvex function, it may be the case that there exist nonconvex problems where $1/t$ implies exponentially slow convergence.

		Moreover, assuming that a function has bounded stochastic gradients, \eg   $\expect\|f_{s_{t}} (x_t)\|\le M$, it is also easy to show that after $T$ steps each with stepsize $\gamma_t = c/t$, then the distance of the current SGD model $x_T$ from the initial iterate $x_0$ satisfies
		$$\mathbb{E}\|x_T-x_0\| \le 2 M \log(T).$$
		This implies that if the optimal model is at distance $\Omega(M d)$ from $x_0$, we would require at least $O(e^{M\cdot d})$ iterations to reach it in expectation.

\end{document}